\documentclass[preprint,12pt]{elsarticle}
\usepackage{pgfplots} 

\usepackage{amssymb}
\usepackage{graphicx}
\usepackage{amsmath}
\usepackage{amssymb}
\usepackage{booktabs}
\usepackage{amsmath}
\usepackage{amssymb}
\usepackage{booktabs}
\usepackage{multirow}
\usepackage{array}
 \usepackage{amsmath}\usepackage{fontawesome5}
\newcommand{\cirbd}{\mathrel{\text{\faDotCircle[regular]}}}
\usepackage{dsfont}
\usepackage[rightcaption]{sidecap}
 \usepackage{tikz}
\usepackage{amsmath}
\usepackage{graphicx}
\usepackage{booktabs}
\usepackage{multirow}
\usepackage{caption}
\usepackage{lmodern}
\usepackage[pagebackref,breaklinks,colorlinks]{hyperref}
\usepackage{times}
\usepackage{epsfig}
\usepackage{graphicx}
\usepackage{amsmath}
\usepackage{amssymb}
\usepackage{adjustbox}
\usepackage{tabularx}
 \usepackage{verbatim}
\usepackage{bm}
\usepackage{subcaption}

 \usepackage{algorithm}

\usepackage{algorithmic}%

\usepackage[capitalize]{cleveref}
\crefname{section}{Sec.}{Secs.}
\Crefname{section}{Section}{Sections}
\Crefname{table}{Table}{Tables}
\crefname{table}{Tab.}{Tabs.}

\journal{Artificial Intelligence}

\begin{document}

\begin{frontmatter}



\title{Relating Events and Frames Based on Self-Supervised Learning and Uncorrelated Conditioning for Unsupervised Domain Adaptation}


\author{Mohammad Rostami, Dayuan Jian, and Ruitong Sun}

\affiliation{organization={University of Southern California}
}
\begin{abstract}
Event-based cameras provide accurate and high temporal resolution measurements for performing computer vision tasks in challenging scenarios, such as high-dynamic range environments and fast-motion maneuvers. Despite their advantages, utilizing deep learning for event-based vision encounters a significant obstacle due to the scarcity of annotated data caused by the relatively recent emergence of event-based cameras. To overcome this limitation, leveraging the knowledge available from annotated data obtained with conventional frame-based cameras presents an effective solution based on unsupervised domain adaptation. We propose a new  algorithm tailored for adapting a deep neural network trained on annotated frame-based data to generalize well on event-based unannotated data. Our approach incorporates uncorrelated conditioning and self-supervised learning in an adversarial learning scheme to close the gap between the two source and target domains. By applying self-supervised  learning, the algorithm learns to align the representations of event-based data with those from frame-based camera data, thereby facilitating knowledge transfer.
Furthermore, the inclusion of  uncorrelated conditioning ensures that the adapted model effectively distinguishes between event-based and conventional data, enhancing its ability to classify event-based images accurately.
Through empirical experimentation and evaluation, we   demonstrate  that our algorithm surpasses existing approaches designed for the same purpose using two   benchmarks. The superior performance of our solution is attributed to its ability to effectively utilize annotated data from frame-based cameras and transfer the acquired knowledge to the event-based vision domain.
\footnote{This paper is based on results partially presented at the 2023 International Conference on Computer Vision  \cite{jian2023unsupervised}.}
\end{abstract}





\begin{keyword}
 unsupervised domain adaptation, event-based cameras,   self-supervised learning 
\end{keyword}

\end{frontmatter}



\section{Introduction}
\label{sec:intro}

Inspired by biological visual systems, event-based cameras are specifically designed to measure image pixels independently and asynchronously by recording changes in brightness relative to a reference level~\cite{lichtsteiner2008128,gallego2020event,posch2014retinomorphic}. These cameras produce a continuous stream of events that encode essential information, such as the location, time, and polarity of intensity changes. Compared to traditional frame-based cameras, event-based cameras offer remarkable advantages, including high dynamic range, high temporal resolution, and low latency, making them highly suitable imaging devices for performing computer vision (CV) tasks that involve high dynamic range scenes and fast-moving objects~\cite{barrios2018movement,alonso2019ev}.
Despite the promising potential that event-based cameras offer, integrating artificial intelligence, particularly deep learning, for automating computer vision tasks in this domain is challenging. Despite similarities between deep learning and the nervous system~\cite{morgenstern2014properties}, deep learning is not as efficient from several perspectives. A major hurdle stems from the fact that training deep neural networks heavily relies on large datasets that are manually annotated~\cite{rostami2018crowdsourcing}. However, the availability of annotated event-based data is scarce due to the relatively recent emergence of these cameras. In fact, event-based data constitutes only a mere $3.14\%$ of the existing vision data~\cite{hu2020learning}. This ratio is  not going to change in the short future because data annotation in general is a time-consuming task that must be performed manually.
This scarcity of annotated data   poses a considerable limitation because deep neural networks generally require large labeled dataset to learn complex representations and generalize well during testing.  

To address the data scarcity challenge and broaden the applicability of deep learning in event-based vision, we can rely ideas from the transfer learning literature~\cite{maqueda2018event} to benefit from existing annotated frame-based data. Specifically, unsupervised domain adaptation (UDA) can be used to leverage the knowledge that can be obtained from learning a source domain with annotated data, e.g., frame-based data or synthetically generated data,  to train a model for a related target domain with limited annotated data, e.g., event-based data~\cite{Messikommer20ral,zhao2022transformer,planamente2021da4event}. 
 The fundamental approach to implement UDA is to reduce the distributional domain gap between the two domains by mapping data from both domains into a shared  embedding space \cite{baktashmotlagh2013unsupervised,bousmalis2017unsupervised,tzeng2017adversarial,pan2019transferrable,rostami2021transfer,rostami2022increasing,rostami2021cognitively,zhao2022transformer,planamente2021da4event,rostami2020generative}, where the distributions of both domains are aligned and similar. Since the two domains become similar in the shared embedding, a model trained on the source domain   will generalize well in the target domain, relaxing the need for having annotated data on the target domain.

While event-based cameras and frame-based cameras differ significantly in their sensing and data representation, the real-world relationship between their outputs establishes a meaningful connection in terms of information content, i.e., both imaging devices in the end are used as sensory devices of the same physical word. Consequently, a labeled frame-based dataset can serve as a source domain to facilitate training for an event-based domain as the target domain within the context of a UDA problem. Even if the related frame-based data is unannotated, annotating frame-based data is comparatively simpler and less labor-intensive.
Although the data representation and sensing mechanisms are distinct in frame-based and event-based datasets, the underlying visual scenes and objects they capture are related. These relationships allow for the effective transfer of knowledge from the annotated frame-based source domain to the unannotated event-based target domain.  
By aligning the representations of data from both domains into a shared embedding space during the UDA process, the model learns to generalize well across domains. As a result, the learned knowledge from the frame-based domain can improve the performance of the model in the event-based domain.

  Initial works that benefit from UDA in event-based domains consider settings where events and frames were paired, meaning that the pixel-level recordings capture  the same input in both formats~\cite{hu2020learning,zhu2021eventgan}. While effective, these methods face a major limitation because they assumed that both measurement   were performed simultaneously across the domains. As a result, the major benefit is that data annotation can be performed in the frame-based domain which is helpful but necessitates using both imaging devices.
Recent UDA algorithms for event-based data have addressed the challenge of unpaired event-based data using multiple general approaches~\cite{rebecq2019high,Messikommer20ral}. These   approaches go beyond the constraint of paired data and open up new possibilities for leveraging unpaired event-based data for domain adaptation and benefiting from existing annotated frame-based domains.
One effective strategy in this context  draws inspiration from the concept of translation in UDA~\cite{murez2018image} and uses it to generate events that correspond to the given video data~\cite{gehrig2020video,hu2021v2e}. By effectively ``translating'' the video data into corresponding event-based representations, these algorithms enable using the frame-based data for training the event-based domain model.
Another alternative approach involves generating synthetic event-based annotated data~\cite{rebecq2018esim} and utilizing it as the source domain~\cite{planamente2021da4event}. By synthetically creating annotated event-based data, these methods effectively overcome the scarcity of annotated event-based datasets and facilitate the UDA process.
The utilization of unpaired event-based data and the generation of synthetic event-based annotations provide promising solutions to augmenting the event-based training datasets and improving the domain adaptation process.

In this study, we develop a new UDA algorithm  for event-based problems using unpaired frame-based data. Our contribution consists of two main components, each aimed at enhancing the model's generalizability and performance.
Firstly, we leverage self-supervised learning~\cite{doersch2017multi,zhai2019s4l} in combination with data augmentation to improve the model's ability to extract informative features. Self-supervised learning enables the extraction of effective representations   by leveraging the inherent relationships between unannotated data samples. The underlying idea is to project different augmentation variations of an object into a shared latent representation and train an encoder to preserve the identities of all these augmentations. By doing so, the model learns to capture essential features and characteristics of the objects while being robust to variations introduced by data augmentation. This approach enables the model to effectively learn from unpaired frame-based data and generalize its knowledge to event-based data.
Secondly, we introduce a novel uncorrelated conditioning loss term to further regularize the model's learning process. This additional loss term provides the model with supplementary information, ensuring that the latent vector representation of an object under event-based cameras remains uncorrelated with how the object appears under event cameras. By enforcing this constraint, the model is encouraged to focus on capturing the unique information and dynamics of events, distinct from the appearance-based information present in frame-based data. This regularization enhances the model's ability to effectively adapt to the event-based domain.

In our experiments, we demonstrate the effectiveness of our UDA algorithm by evaluating its performance on two benchmark datasets. The results show considerable performance improvements over state-of-the-art methods. Specifically, we observe a remarkable 2.0\% performance improvement on the Caltech101 $\rightarrow$ N-Caltech101 domain adaptation task and a   3.7\% improvement on the CIFAR10 $\rightarrow$ CIFAR10-DVS domain adaptation task. We also offer ablative experiments to demonstrate that both our ideas are necessary for an optimal performance and provide analytic experiments to offer insights about our algorithm. 
Our findings highlight the potential of our proposed UDA algorithm to learn a more robust   representation of events in an unsupervised manner. By effectively leveraging unpaired frame-based data and incorporating self-supervised  learning and uncorrelated conditioning, one can successfully adapt models for  event-based domains.  

The rest of the paper is as follows. In Section 2, we review relevant work to explain our work in the context of existing works. In Section 3, we provide the problem formulation. Section 4 is devoted to our proposed algorithm. We provide extensive empirical and analytic experiments is Section 5. Finally, Section 6 concludes the paper with a brief future research statement.

\section{Related Work}

\label{sec:Related}
\subsection{Unsupervised Domain Adaptation for Frame-Based Tasks}

When dealing with frame-based measurements from the same frame-based modality, e.g. electro-optical modality, UDA has been explored extensively in the literature. \cite{wilson2020survey,wang2018deep}.
The core principle behind UDA in frame-based tasks involves training an encoder that couples the two domains at its output by aligning the corresponding probability distributions. The shared embedding space at the output of the encoder allows for a seamless transfer of knowledge between the two domains.  There are two prominent approaches to train the shared encoder: generative adversarial learning~\cite{tzeng2017adversarial,long2018conditional,zhang2018collaborative,chen2020adversarial} and distribution alignment using suitable loss functions~\cite{long2017deep,venkateswara2017deep,stan2022secure,liang2019exploring,le2019deep,yang2018learning,gabourie2019learning,stan2022domain,rostami2021lifelong,rostami2023domain}.
In methods based on   adversarial learning, the generator subnetwork within the architecture of a generative adversarial network (GAN) \cite{goodfellow2020generative} serves as the shared encoder. This generator subnetwork is trained to compete against a discriminative subnetwork with the objective of learning a domain-agnostic yet discriminative embedding space. The ultimate goal is to align the distributions of data from the source and target domains indirectly. This alignment occurs at the output of the generative subnetwork when the discriminative subnetwork becomes incapable of distinguishing between samples from these two domains. The adversarial learning process is achieved through the use of a min-max objective function, which is solved through alternation and governs the adversarial training process.

In the second approach, the main focus is on quantifying the dissimilarity between the empirical distributions of two domains, as they emerge from the shared encoder \cite{ghifary2016deep,morerio2017minimal,damodaran2018deepjdot,lee2019sliced,stan2021unsupervised,rostami2019deep}. The pivotal question is how to identify a suitable metric for this purpose. A classic metric  for this purpose the Maximum Mean Discrepancy (MMD) \cite{long2015learning,long2017deep} which is employed for aligning the means of two distributions. An improvement over MMD is to  align  distribution correlations \cite{sun2016deep}, thus incorporating second-order statistics.
Although aligning lower order probability moments for domain alignment is straightforward, it disregards potential mismatches in higher moments. To address this limitation, the Wasserstein distance (WD) has gained traction in capturing information from higher-order statistics \cite{courty2017optimal,damodaran2018deepjdot}. Notably, Damodaran et al. \cite{damodaran2018deepjdot} showcased the superiority of using WD in Unsupervised Domain Adaptation (UDA) when compared to MMD or correlation alignment \cite{long2015learning,sun2016deep}. Furthermore, there exist theoretical guarantees that demonstrate domain alignment using WD will minimize an upper bound for the expected error in the target domain \cite{redko2017theoretical}.
Another viable metric utilized for this purpose is the sliced Wasserstein distance \cite{stan2021unsupervised,stan2022secure,rostami2019deep,rostami2023overcoming,rostami2021lifelongww}. The sliced Wasserstein metric offers several advantages over optimal transport. It is more scalable, capable of handling larger datasets and higher-dimensional spaces, addressing a limitation of optimal transport. Despite working with one-dimensional projections, the sliced Wasserstein distance provides a good approximation of the optimal transport distance, capturing important information about data distributions. Moreover, its flexibility allows practitioners to adapt the number and directions of slices, providing a customizable approach to match the available computational resources.   Once the appropriate metric is selected and the loss function is formulated, the encoder is trained to directly minimize the loss function, effectively narrowing the distribution gap between the two domains. SWD and MMD have been used concurrently to benefit from the strengths of both metrics~\cite{wu2023unsupervised}.

\subsection{Unsupervised Domain Adaptation for Event-Based Tasks}

While dealing with supervised learning problems in event-based domains, one of the major hurdles is the scarcity of annotated data, which poses a common challenge~\cite{gouda2023improving}. Unlike frame-based tasks, there has been relatively limited research focused on utilizing the UDA framework to address tasks involving unannotated event-based data. The main reason for this disparity is that we can mostly rely on frame-based data as the source domain for adaptation. This limitations exists because most existing annotated datasets are frame-based and unlike frame-based domains, we don't have many existing annotated event-based datasets. Unfortunately, when attempting to align the distributions between frame-based and event-based domains, the gap between them can be considerably larger compared to the domain gap between two domains belonging to the same frame-based modality.
The consequence of this  distribution gap is that simply using a frame-based UDA methods to align the source and target distributions may not suffice to effectively benefit from UDA. More sophisticated and innovative approaches are required to bridge this   discrepancy and enable successful adaptation in event-based domains with limited annotated data.  

Drawing inspiration from image-to-image translation in frame-based UDA \cite{murez2018image},   generative models can be used to map images from one domain to the other domain, effectively aligning the distributions in the input image space of one of the two domains. For example, by translating target domain images to the source domain, it becomes possible to directly utilize a classifier trained on the source domain. As a result, some existing studies have adopted adversarial training techniques to translate event-based data into frame-based images\cite{wang2019event, wang2020eventsr, choi2020learning}.
However, a notable drawback  is   reliance on paired image-event data, where corresponding pairs of images and events must be measured on the same scene, to learn frames to the corresponding events. Establishing such cross-domain correspondences can be challenging, especially when generating paired data becomes impractical due to factors like different lighting conditions or other variations between the domains. This limitation impedes the application of these techniques in real-world scenarios where obtaining precisely matched data is difficult.
To overcome this limitation, a group of works attempt to separate domain-agnostic features (such as geometric structures) from domain-specific features (such as texture)~\cite{zhang2020learning}. By disentangling these aspects, the   aim is to find a common representation that is shared between the event-based and the frame-based domains, facilitating a more effective alignment of their distributions. Hence, the need for precisely paired data is alleviated, allowing for a more practical and scalable adaptation.

Another alternative method is utilization of generative models with appropriate architectures, specifically video-to-event translation. The primary goal of this approach is to transform videos consisting of frames into synthetic event-based data, enabling direct training of the model within the event space. Two main techniques can be employed to achieve this: model-based translation and data-driven translation.
Model-based translation \cite{rebecq2019high, gehrig2020video, hu2021v2e} involves the use of specific models designed to perform the translation from video to events. These models are tailored to learn the underlying patterns and correlations between video frames and corresponding events, facilitating the generation of accurate and meaningful event-based representations. 
On the other hand, data-driven translation  \cite{zhu2021eventgan} leverages large-scale datasets to train generative models. These models learn from the data distribution and can effectively convert frame-based videos into event-based representations. Data-driven approaches are flexible and capable of capturing complex relationships between the input and output domains. However, similar to model-based translation, their limitation lies in being confined to video-to-event translation tasks.

The   most similar to the current research divides the embedding space into shared and sensor-specific features and employs an event generation model to align  both domains~\cite{Messikommer20ral}. Despite high-level similarities, our work takes a different approach, aiming to enhance model generalizability by incorporating self-supervised  learning. Additionally, we introduce a novel loss term to ensure that object representations become uncorrelated with their event-based measurements. Our experiments showcase that these   contributions empower us to achieve the performance level of supervised learning through UDA, consequently reducing the necessity for data annotation in certain scenarios. This methodology becomes particularly promising when dealing with limited annotated event-based data, thus offering valuable insights to advance UDA techniques in tasks involving event-based data.

\textcolor{black}{Finally, we note recent advances in domain adaptation have extended beyond classic settings to address more complex scenarios such as partial and heterogeneous domain adaptation. In Partial Domain Adaptation (PDA), where the target domain contains only a subset of the source classes \cite{cao2018partial,tian2021partial,cao2019learning}, techniques like selective alignment \cite{li2020dual} and critical class discovery \cite{li2022critical}   have been proposed to prevent negative transfer from irrelevant source categories. While PDA is not the focus of our work, the notion of selectively aligning informative structures remains conceptually aligned with our use of self-supervised constraints.
Heterogeneous Domain Adaptation (HDA) addresses scenarios where the feature spaces or label spaces of source and target domains differ \cite{liu2020heterogeneous,duan2012learning,fang2022semi}. For example, semantic correlation transfer methods \cite{zhao2022semantic} have been developed to map semantic structures across modalities, which is conceptually relevant to our setting where frame and event data represent different sensing modalities. Our use of uncorrelated conditioning and adversarial alignment can be viewed as implicitly learning such semantic bridges without requiring paired or labeled event data.
To the best of our knowledge, prior works on UDA for event-based vision \cite{rebecq2019high} do not incorporate either self-supervised regularization or decorrelation of sensor-specific features. Our proposed method differs by addressing the heterogeneous domain challenge explicitly through architectural and loss design choices, allowing generalization from frames to events even under large distributional and representational gaps.
}

\section{Problem Formulation: Domain Adaptation from Frames to Events}
\label{sec:Description}
\begin{figure}[h]
    \centering       
    \includegraphics[width=13.6cm, height=4.1cm]{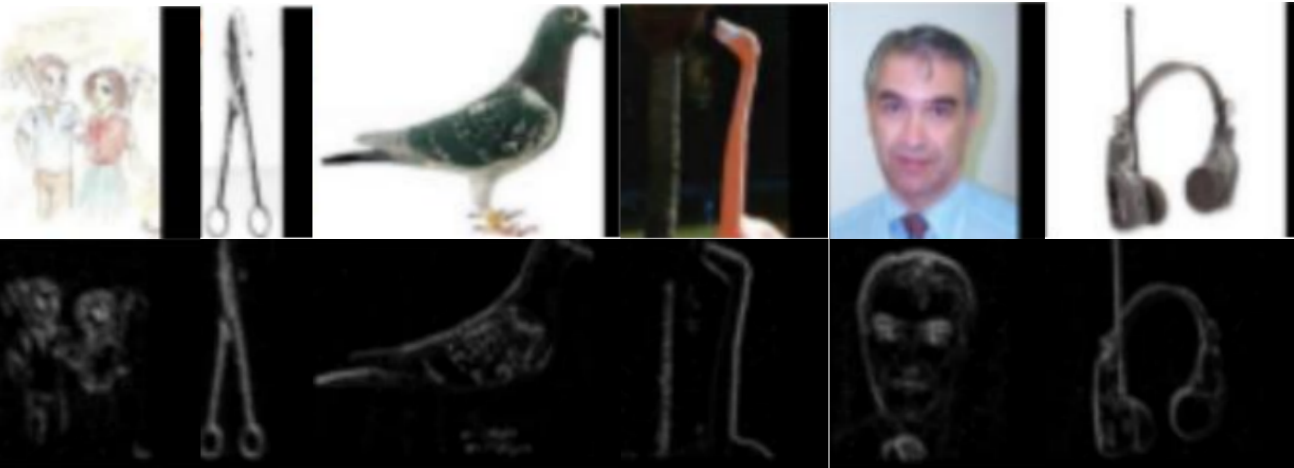}
 \caption{Randomly drawn frame (row 1, Caltech101\cite{https://doi.org/10.48550/arxiv.1604.01518}) and event (row 2, N-Caltech101\cite{https://doi.org/10.48550/arxiv.1507.07629})  pairs: we observe larger domain gaps compared to two frame-based domains. However, we also observe that some high-level features such as edges are shared across the two domains.}
    \label{fig1}
\end{figure}

Our   objective is to develop a a UDA algorithm to  train a model using  labeled RGB images (frames) and subsequently adapt it to generalize well om events data using solely unannotated event data, relaxing the need for  data annotation. Importantly, we  assume that we do not have access to paired frames and event data which is the more practical scenario. Compared to most UDA problems, we need to address a larger domain gap in this setting.
In Figure~\ref{fig1}, we have visualized  samples of paired frame and event data from two commonly used event-based datasets, i.e., Caltech101 frame-based dataset \cite{https://doi.org/10.48550/arxiv.1604.01518})  N-Caltech101 event-based dataset \cite{https://doi.org/10.48550/arxiv.1507.07629}. As it can seen, it is evident that these pairs exhibit significantly more differences compared to the benchmarks typically utilized in frame-based UDA, e.g., the Office-Home~\cite{venkateswara2017deep} or DomainNet~\cite{peng2019moment} datasets.
The reason is that data modalities are significantly different between the event-based and the frame-based cameras. 
As a result, conducting UDA for tasks involving   event-based and frame-based data as the target and the source domains is more challenging due to larger distributional gaps. However, it is worth noting that event and frame-based cameras capture scenes in the physical world, and they often share substantial information overlap, which makes knowledge transfer between these two domains a feasible proposition. For example, we observe in Figure~\ref{fig1} that some edges are shared across image pairs.
To close the distributional gaps between the two domains, we   explore the possibility of aligning the content representations obtained from events with those derived from frames in a latent space. By achieving such alignment, we  enable the utilization of a frame-trained classifier model on the events data because the two domains are indistinguishable in the latent space.

To formulate our problem, consider that we have access to   unannotated event data instances $\bm{y}_e$ from the event domain $\mathcal{Y}_e$ and annotated frame-based image data $\bm{y}_f$ from the frame domain $\mathcal{Y}_f$. Our goal is to establish a mapping between these data points in a shared domain-agnostic embedding space   $\mathcal{Z}$. We refer to this shared embedding space as the content space because its purpose is to capture high-level shared information content across the two domains.
To achieve this goal, we employ two distinct domain-specific deep encoders $E_f(\cdot)$ and $E_{e,cont}(\cdot)$. These two encoders model the content space as their output space.  
The encoder $E_f(\cdot)$ projects frames into the content feature space, and similarly, the encoder $E_{e,cont}(\cdot)$ projects events into the content feature space at its output. The reason behind using separate encoders as opposed to many UDA setting is the existence of  large domain gap between events and frames.

Additionally, we utilize a classifier network denoted as $C(\cdot)$, which takes the extracted content features from the output-space of the encoders and maps them to the label space. Given that we possess labeled frames, we can easily employ standard supervised learning techniques to train the combined end-to-end neural network $C(E_f(\bm{y}_f))$. This process involves solving an     empirical risk minimization (ERM) problem using a suitable loss, allowing us to predict frame labels effectively. The end-to-end network can be trained using only the frame data  
but we would like to train the encoders such that we learn a shared content representation for both frames and events in the common embedding space $\mathcal{Z}$ despite the differences between the two domains at the input-space of the encoders. By doing so, we aim to leverage the labeled frame data to train the classifier subnetwork and eventually make predictions on event data without requiring explicit event labels because the gap between the two domains is minimal in the content domain by achieving successful cross-domain knowledge transfer.

More specifically, our goal is to align the content representations of events    $E_{e,cont}(\bm{y}_e)$  with the content representations of frames $E_f(\bm{y}_f)$. The straightforward scenario would be having access to paired event-frame data. In such a case, we could   employ a point-wise loss function to align these paired data points in the content space, effectively extracting the shared content representation knowing that the pair describe the same scene.
However, in our more practical setting, we do not have direct access to frame-event pairs. To address this challenge, we  develop  our algorithm based on adversarial machine learning and model the encoders as generator subnetworks in  adversarial learning. 
Once we achieve domain alignment and have successfully trained the classifier subnetwork, we can utilize the end-to-end combined model $C(E_{e,cont}(\bm{y}_e))$ to     effectively make predictions on the event data using the shared classifier subnetwork.

\section{Proposed Unsupervised Domain Adaptation Algorithm}
\label{sec:Method}

\textcolor{black}{In this section, we provide the details for  our proposed   algorithm designed to transfer knowledge from labeled frame-based data to unlabeled event-based data. The core idea is to learn domain-agnostic content representations using adversarial training while enhancing alignment with two novel regularization techniques: self-supervised learning and uncorrelated conditioning.
We begin by outlining the high-level structure of our framework and then describe each component,  followed by the training objective in detail.}

\subsection{Overview of Architecture }

\textcolor{black}{Our architecture consists of three main modules:
\begin{itemize}
    \item Encoders: To extract content and attribute features from frame and event data.
    \item Generators and Discriminators: To generate synthetic events from frame content and align feature distributions via adversarial training.
    \item Classifiers: To perform label prediction based on shared content features.
\end{itemize}}

\begin{figure}[ht]
    \centering
    \includegraphics[width=10cm, height=10.77 cm]{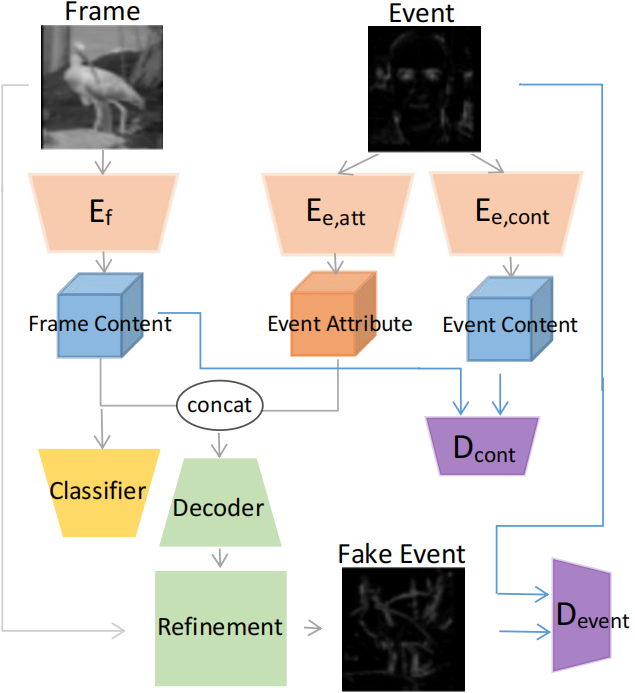}
    \caption{\small Architecture for our method during training.
    During testing, only event content encoder and classifier are needed.}
    \label{fig2}
\end{figure}

The architecture of our framework is visually depicted in Figure \ref{fig2}.
The core idea is to use adversarial training to generate synthetic event data from the content features of frame images, with the aim of making these synthetic events indistinguishable from real events. This adversarial learning process indirectly aligns the content features of both domains.
In Figure \ref{fig2}, we have an image encoder and an event encoder, both responsible for generating features that represent the content within images and events, respectively. Importantly, the output space of these two encoders shares the same dimension. The content features obtained from these encoders are then fed into a classifier, enabling label predictions.

The primary objective in our architecture in Figure \ref{fig2}  is to train the encoders in such a way that the content representations of event data exhibit the same distribution as the content representations of frame data in the shared embedding space $\mathcal{Z}$. To further enhance the process, we introduce a second event encoder denoted as $E_{e,att}(\cdot)$, which generates feature attributes corresponding to the visual characteristics of objects observed under an event camera.
The frame content and the event attributes are then passed through a decoder, generating synthetic events. However, to ensure the quality of these synthetic events, we route them through a refinement network, which further improves the fidelity of the fake event data. The ultimate goal of this process is to generate high-quality fake events that effectively encapsulate the content within the frame and accurately represent the content as it would appear under an event camera.
The crucial aspect of our approach is that if the synthetic events look realistic and are difficult to distinguish from real events, it indicates that the shared content representations indeed encode valuable and shared information from both events and frames. This alignment and generation process using adversarial learning allows us to effectively transfer knowledge between the two domains, facilitating successful UDA for event-based tasks.

\subsection{Encoders and Content Space}

To complete the necessary components  for an adversarial learning architecture, we introduce a content discriminator network denoted as $D_{cont}(\cdot)$. This discriminator is responsible for distinguishing whether a given content representation originates from an event or an image. By training this network against both encoders, we can effectively learn domain-agnostic content features that bridge the gap between events and frames.
To impose additional constraints and enhance the alignment process, we draw inspiration from the concept used in cycleGAN \cite{CycleGAN2017}. Using cycleGAN particularly helps to overcome the challenge of having unpaired data. For this purpose,  we employ a generated fake event and pass it through the two event encoders, obtaining its event attribute and event content. Then, we compare these attributes and content with those generated by the original fake event using the $\ell_1$ loss function. This constraint further enforces the alignment between the content features of events and frames.
In our pursuit of generating realistic fake events, we introduce a second discriminator network, $D_{event}(\cdot)$, which serves to distinguish between real and fake events. The fake event, being generated from content features with known labels, has its content feature sent to the classifier for end-to-end training.
The choice of exact architectures for the encoders and the classifier is a design decision and can be tailored based on specific requirements. For simplicity, we   opt for using similar architectures for all three encoders.

\subsection{Loss Functions and Training Objective}

To implement the aforementioned constraints and ensure effective domain alignment, we define specific loss terms and formulate the underlying cumulative optimization problem based on the architecture presented in Figure \ref{fig2}. By iteratively solving the optimization problem that we design and fine-tuning the network, we can achieve successful adversarial learning, generating high-quality synthetic events, and establishing domain-agnostic content features that effectively represent shared information across events and frames.
Because domain gap is large in this problem, training the architecture of Figure \ref{fig2} is a challenging task and simply using GAN losses will not be sufficient. We benefit from   the following loss terms which have been used in the GAN-based UDA literature for domain alignment:

\begin{itemize}
 \item     \textbf{Classification loss on frames, $\mathcal{L}_{cls,frame}$}: it is a
Cross Entropy loss function to implement supervised learning.

 \item  \textbf{Classification loss on fake events, $\mathcal{L}_{cls,fake}$}, it is  Cross Entropy loss  on  fake events to train a model that generalizes better on  events as well.
 
 \item  \textbf{Decoder output loss, $\mathcal{L}_{decoder}$}: the loss consists of two terms. The first term   approximates the event version of a frame image according to the $\ell_1$ loss  \cite{Gallego2015EventbasedCP}.  The second term enforces  cycle consistency \cite{CycleGAN2017} to benefit from unpaired data. The decoded outputs of event attribute of fake events and frame contents are enforced to be  close to the decoded output of event attributed of real event and frame content.  
 \begin{equation}
\begin{split}
\mathcal{L}_{decoder}=& ||D_{cont}(E_f(\bm{y}_f))-\bm{y}_f||_1 +  || fake-D_{cont}(E_{e,cont}(fake))||_1\\&+  || \bm{y}_e-D_{cont}(E_{e,cont}(\bm{y}_e))||_1. 
\end{split}
\label{eq1}
\end{equation}

\item \textbf{Cycle loss on feature representations of contents and attributes:} $\mathcal{L}_{cyc,cont}$ and $\mathcal{L}_{cyc,att}$ 
These are $\ell_1$ losses imposed on the features used to generate the fake events and the features extracted from them:
\begin{equation}
\begin{split}
&\mathcal{L}_{cyc,cont} = ||E_f(\bm{y}_f)-E_{e,cont}(fake)||_1\\&\mathcal{L}_{cyc,att} = ||E_{e,att}(\bm{y}_e)-E_{e,att}(fake)||_1.
\end{split}
\end{equation}
Both loss terms help to make the embedding space discriminative.

\item \textbf{Content Discriminator:} There is a discriminator network to differentiate frame content and event content. The purpose of this discriminator is to make frame content features and event content features have the same distribution, denoted  as $g(\cdot)$ = \(\log(sigmoid(\cdot))\) and $h(\cdot)$ = \(\log(1-sigmoid(\cdot))\). We use relativistic average discriminator\cite{jolicoeurmartineau2018relativistic} loss as follows:
\begin{equation}
\begin{split}
\mathcal{L}_{dis,cont} =   \frac{1}{N}\sum 
& \{\bm{g}[D_{cont}(E_f(\bm{y}_f)) -\frac{1}{N}\sum D_{cont}(E_{e,cont}(\bm{y}_e))] +\\& \bm{h}[D_{cont}(E_{e,cont}(\bm{y}_e))-\frac{1}{N}\sum D_{cont}(E_f(\bm{y}_f))]\}
\end{split}
\end{equation}

\item \textbf{Generative Loss for encoder $E_f, E_{e,cont}$:} these loss terms enforce the adversarial competition to train the encoders:
\begin{equation}
\begin{split}
\mathcal{L}_{encoder,cont} =   \frac{1}{N}\sum  &\{\bm{h}[D_{cont}(E_f(\bm{y}_f)) -\frac{1}{N}\sum D_{cont}(E_{e,cont}(\bm{y}_e))] + \\&\bm{g}[D_{cont}(E_{e,cont}(\bm{y}_e))-\frac{1}{N}\sum D_{cont}(E_f(\bm{y}_f))]\}
\end{split}
\end{equation}

\item \textbf{Event Discriminator loss:} We have a discriminator subnetwork to differentiate fake events and real events.   We use a relativistic average discriminator \cite{jolicoeurmartineau2018relativistic} to train this subnetwork:
\begin{equation}
\begin{split}
 \mathcal{L}_{dis,e} = 
 \frac{1}{N}\sum & \{\bm{g}[D_{e}(\bm{y}_e) \\&-\frac{1}{N}\sum D_{e}(fake)]  \bm{h}[D_{e}(fake)-\frac{1}{N}\sum D_{e}(\bm{y}_e)]\}
\end{split}
\end{equation}
\item Similarly, the generator loss is defined as: 
\begin{equation}
\begin{split}
\mathcal{L}_{gen,e} =\frac{1}{N}\sum & \{\bm{h}[D_{e}(\bm{y}_e)\\& -\frac{1}{N}\sum D_{e}(fake)] +  \bm{g}[D_{e}(fake)-\frac{1}{N}\sum D_{e}(\bm{y}_e)]\}
\end{split}
\label{eq72}
\end{equation}

\item \textbf{Orthogonal Normalization:} $\mathcal{L}_{orth}$, Inspired by BigGAN \cite{brock2019large},   this regularization term is crucial to stabilize the training in adversarial learning which is known to suffer from instability:
\begin{equation}
\begin{split}
L_{orth} = \beta ||W^TW\cirbd(\mathds{1}-I)||^2_F,
\end{split}
\label{eq7}
\end{equation}
where \(W\) is the weight matrix, \(\mathds{1}\) is the matrix where all elements are 1, \(I\) is the identity matrix, \(\cirbd\) is element-wise multiplication, and denotes $\|\cdot\|_F$ Frobenius norm.

\end{itemize}

In our work, we minimize the sum of the above loss terms for domain alignment~\cite{planamente2021da4event}. Except for the first two loss terms, the rest of the terms constrain the encoders to align the distributions in the content latent space. In our experiments, we observe that using the above loss terms lead to a decent performance, but we propose two novel ideas based on self-supervised  learning and uncorrelated conditioning for further improvement upon this baseline. To the best of our knowledge, these loss terms are novel in the context of UDA. We think that both ideas are adoptable in other UDA problems with large domain gaps.

\subsubsection{Self-Supervised Learning for Improved Content Extraction} 
   
\begin{figure}[h]
    \centering
    \includegraphics[width=9cm, height=7.46cm]{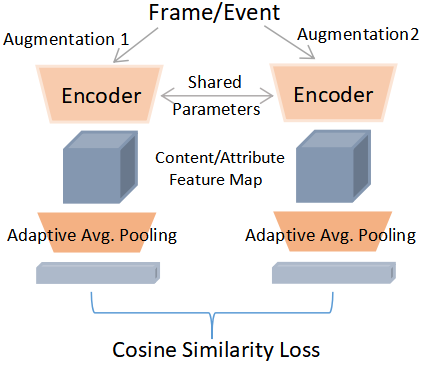}
    \caption{The pipeline for computing the self-supervised learning loss: the loss term  is computed for all three encoders using the augmented versions of data and the original version.}
    \label{fig3}
\end{figure}

   We use  self-supervised learning   to effectively learn meaningful representations from unlabeled data in the content space. 
   In Figure ~\ref{fig3}, we present our pipeline for calculating the self-supervised learning loss term. In this process, we utilize data augmentation to create diverse versions of a frame image and ensure that all these variations share identical feature representations in the latent space  \cite{https://doi.org/10.48550/arxiv.2006.07733,chen2020big,chen2020mocov2}. We add the self-supervised learning loss as an additional constraint to enforce convergence to a  solution that extracts more generalizable features.   Let \(f(\cdot)\) denote an encoder and an average pooling layers. Let \(f(\bm{y})\) and \(f(\bm{y}')\) will be two feature vector representations of a single input \(\bm{y}\) that are produced using augmented versions. We enforce a hard thresholding constraint:
\begin{equation}
 \frac{\langle f(\bm{y}),f(\bm{y}')\rangle}{||f(\bm{y})||_2\cdot||f(\bm{y}')||_2} ==1, 
\label{eq112}
\end{equation}
where $\langle\cdot,\cdot\rangle$ denotes vector inner product and $||\cdot||_2$ is  $l_2$-norm. In other words, Eq.~\eqref{eq112} enforces the augmented versions of a single input share the same embedding feature at the output of an encoder.  We enforce this constraint on output-space of all the three encoders for data from both domains.

  \subsubsection{Uncorrelated Conditioning}

In many computer vision tasks, the primary focus is often on the objects captured in an image   to perform the task. The idea behind using the uncorrelated conditioning loss function is to leverage the entire information content present in an event image. Besides objects, such as a pedestrian, an image contains other valuable information that we can leverage. For example, we can determine whether the image is blurred or clear, whether it was taken during the day or at night, whether it was captured by a high-resolution or low-resolution camera, and whether it originated from a frame camera or an event camera. These high-level properties can be regarded as additional features that we can extract from images.
The key idea that we explore is that these properties or features are uncorrelated with each other. For instance, knowing whether an image is blurred or not does not provide any information about the presence of a pedestrian in the image. Similarly, knowing whether the image was taken during the day or night does not reveal anything about the quality of the camera used.
For tasks such as image classification, object detection, or image segmentation, our approach aims to train the model to extract not only features related to objects within an image but also domain-specific features. Our approach necessitates extracting features from event data as well because objects captured by event cameras have distinct appearances compared to those captured by frame cameras. Hence, we generate synthetic or translated event images to account for these differences.
To address the fact that frame-based features and event-based camera features are uncorrelated, we impose an uncorrelated conditioning constraint. This constraint ensures that information about an object's nature in the frame image does not provide any indication of how the corresponding event image would appear. Likewise, the appearance of an event image reveals no information about the content of the event captured. In other words, the uncorrelated conditioning loss helps to mitigate the effect of negative transfer across two domains by making the domain-specific features uncorrelated.

\begin{figure}[t]
    \centering
    \includegraphics[width=9cm, height=7.46cm]{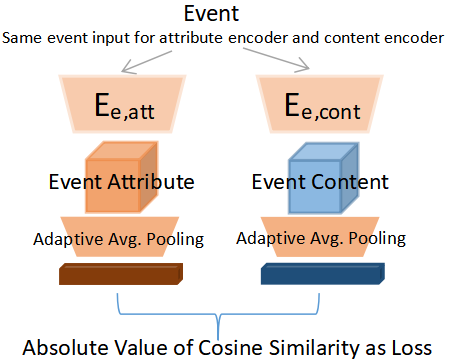}
    \caption{The pipeline for computing the uncorrelated conditioning loss: we decorrelate the event attributes and the corresponding content by making the corresponding feature vectors dissimilar.}
    \label{fig41}
\end{figure}

In Figure ~\ref{fig41}, we present the pipeline for calculating the uncorrelated conditioning loss term using the event data. We use the content encoder and the attributed encoder of event data to compute the uncorrelated conditioning loss:
\begin{equation}
 \frac{\langle f_{e,a}(\bm{y}_e),f_{e,c}(\bm{y}_e)\rangle}{||f_{e,a}(\bm{y}_e)||_2\cdot||f_{e,c}(\bm{y}_e)||_2}==0,
\label{eq8}
\end{equation}
where \(f_{e,a}(\bm{y}_e)\) and \(f_{e,c}(\bm{y}_e)\) are the two feature vector representations of attribute and content, respectively, for the same event \(\bm{y}_e\). When the dot product of two vectors is zero in Eq.~\eqref{eq8}, they become uncorrelated.
The constraint in Eq.~\ref{eq8} helps to decorrelate 
attribute and content for events and  helps in maintaining the independence between the features extracted from objects and the features specific to the event camera and making the shared content more relevant to the frame-based feature. Hence, this constraint leads to improved knowledge transfer, enabling the model to disentangle the diverse information present in in event images

Following our discussion about constraints \eqref{eq8} and \eqref{eq112}, we   denote  the content vector for frame \(\bm{y}_f\) as \(f_{fr}(\bm{y}_f)\).
 We can then add the following terms to the optimization objective function as   additional regularizing losses to benefit from self-supervised learning and uncorrelated conditioning:

\begin{equation} 
\begin{split}
\min \sum_{y_e,y_f}  
&-\lambda_1\frac{ \langle f_{e,a}(\bm{y}_e),f_{e,a}(\bm{y}_e')\rangle}{||f_{e,a}(\bm{y}_e)||_2\cdot||f_{e,a}(\bm{y}_e')||_2} - 
\lambda_2\frac{ \langle f_{e,c}(\bm{y}_e),f_{e,c}(\bm{y}_e')\rangle}{||f_{e,c}(\bm{y}_e)||_2\cdot||f_{e,c}(\bm{y}_e')||_2} \\& -\lambda_3\frac{ \langle f_{fr}(\bm{y}_f),f_{fr}(\bm{y}_f')\rangle}{||f_{fr}(\bm{y}_f)||_2\cdot||f_{fr}(\bm{y}_f')||_2} 
+ \lambda_4 |\frac{\langle f_{e,a}(\bm{y}_e),f_{e,c}(\bm{y}_e)\rangle}{||f_{e,a}(\bm{y}_e)||_2\cdot||f_{e,c}(\bm{y}_e)||_2}|,
\end{split}
\label{eq17}
\end{equation}
where \(\lambda_1\), \(\lambda_2\), \(\lambda_3\) and \(\lambda_4\)  are trade-offer parameters. Our total loss consists of sum of the all terms listed in Equations \eqref{eq1} through \eqref{eq7} and Eq. \eqref{eq17}.

\textcolor{black}{
To summarize the components of our method, 
Table \ref{tab211} offers a concise overview of all the loss functions used in training, along with their symbolic representations and functional roles. This summary helps disambiguate the purpose of each term and illustrates how they collectively enforce feature alignment, semantic consistency, and domain-invariant representation learning. Additionally,
we provide Algorithm \ref{Algo}  which outlines the training procedure step-by-step and highlights the interplay between different modules and loss terms. This procedural view clarifies how our framework integrates adversarial learning, self-supervised regularization, and uncorrelated conditioning into a unified optimization loop. Each module—encoders, generator, discriminators, and classifier—is updated iteratively based on the respective objective functions.
Together, the algorithm and table serve as a compact reference to the full methodology, providing both a high-level understanding and a practical implementation guide for reproducing and extending our approach.}

\begin{table}[ht]
\centering
\small
\caption{Summary of Loss Functions}
\label{tab211}
\begin{tabular}{lll}
\toprule
\textbf{Loss Name} & \textbf{Symbol} & \textbf{Purpose} \\
\midrule
Frame classification & $\mathcal{L}_{\text{cls,frame}}$ & Supervised loss on labeled frame data \\
Fake event classification & $\mathcal{L}_{\text{cls,fake}}$ & Supervised loss on generated events \\
Content discriminator & $\mathcal{L}_{\text{dis,cont}}$ & Adversarial loss to align content embeddings \\
Content encoder GAN & $\mathcal{L}_{\text{encoder,cont}}$ & Fool the content discriminator \\
Event discriminator & $\mathcal{L}_{\text{dis,e}}$ & Distinguishes real vs fake events \\
Event encoder GAN & $\mathcal{L}_{\text{gen,e}}$ & GAN loss for generating realistic events \\
Decoder loss & $\mathcal{L}_{\text{decoder}}$ & $\ell_1$ reconstruction + cycle consistency \\
Cycle consistency (content) & $\mathcal{L}_{\text{cyc,cont}}$ & Preserve content features \\
Cycle consistency (attribute) & $\mathcal{L}_{\text{cyc,att}}$ & Preserve attribute features \\
Self-supervised loss & $\mathcal{L}_{\text{self-sup}}$ & Regularize with augmentations \\
Uncorrelated conditioning & $\mathcal{L}_{\text{uncorr}}$ & Decorrelate content and attribute features \\
Orthogonal regularization & $\mathcal{L}_{\text{orth}}$ & Stabilize adversarial training \\
\bottomrule
\end{tabular}
\end{table}

\begin{algorithm}[ht]
\caption{Uncorrelated Conditioning}
\label{Algo}
\begin{algorithmic}[1]
\REQUIRE Labeled frame images $y_f$ with labels, unlabeled event data $y_e$
\STATE Initialize: Encoders $E_f$, $E_{e,\text{cont}}$, $E_{e,\text{att}}$; Classifier $C$; Decoder $D$; Discriminators $D_{\text{cont}}$, $D_{\text{event}}$
\FOR{each training iteration}
    \STATE Sample mini-batches of $y_f$ and $y_e$
    \STATE Apply augmentations to $y_f$ and $y_e$ to compute SSL loss $\mathcal{L}_{\text{self-sup}}$
    \STATE $E_f(y_f) \rightarrow \text{content}_f$
    \STATE $E_{e,\text{cont}}(y_e) \rightarrow \text{content}_e$
    \STATE $E_{e,\text{att}}(y_e) \rightarrow \text{attribute}_e$
    \STATE $D(\text{content}_f, \text{attribute}_e) \rightarrow \text{fake\_event}$
    \STATE Compute $\mathcal{L}_{\text{decoder}}, \mathcal{L}_{\text{cyc,cont}}, \mathcal{L}_{\text{cyc,att}}$
    \STATE Compute classification losses: $\mathcal{L}_{\text{cls,frame}}$, $\mathcal{L}_{\text{cls,fake}}$
    \STATE Update $D_{\text{cont}}, D_{\text{event}}$ using adversarial losses: $\mathcal{L}_{\text{dis,cont}}, \mathcal{L}_{\text{dis,e}}$
    \STATE Update $E_f, E_{e,\text{cont}}$ using GAN losses: $\mathcal{L}_{\text{encoder,cont}}, \mathcal{L}_{\text{gen,e}}$
    \STATE Enforce uncorrelated conditioning loss $\mathcal{L}_{\text{uncorr}}$
    \STATE Apply orthogonal regularization loss $\mathcal{L}_{\text{orth}}$
    \STATE Update all network parameters using total loss $\mathcal{L}_{\text{total}}$
\ENDFOR
\end{algorithmic}
\end{algorithm}

\subsection{Theoretical Justifications for Uncorrelated Conditioning}

 \textcolor{black}{
As discussed, uncorrelated conditioning constraint enforces that the domain-specific features  and domain-invariant content features extracted from the same event sample are uncorrelated because it enforces the cosine similarity between them to be zero. This constraint promotes orthogonality between the two feature vectors in the shared representation space. Orthogonality ensures that gradients from one representation do not interfere with the other during backpropagation, facilitating disentangled representation learning. This disentanglement helps mitigate negative transfer by preventing domain-specific noise from contaminating class-relevant content features.
The theoretical justification can be framed within the standard UDA generalization bound developed by Ben-David et al. \cite{ben2006analysis} which relates the generalization error of a model $h$ on a source and the target domain as follows:}

\begin{equation}
\epsilon_T(h) \leq \epsilon_S(h) + \mathcal{D}_{\mathcal{H}\Delta\mathcal{H}}(P_S, P_T) + \lambda,
\end{equation}
 \textcolor{black}{
where $\epsilon_T(h)$ and $\epsilon_S(h)$ are the target domain and source domain risks, $\mathcal{D}_{\mathcal{H}\Delta\mathcal{H}}$ is a metric which measures the discrepancy between the source and target distributions, and $\lambda$ represents the shared error of the ideal joint hypothesis on both domains. In this setting, enforcing uncorrelated conditioning contributes to reducing $\lambda$ by helping the model isolate class-discriminative information that is consistent across domains, hence improving generalization on the target domain when the discrepancy between the two domains is minimized.
In other words, uncorrelated conditioning serves as a structural prior that enhances the transferability of semantic features while suppressing domain-specific interference, leading to more effective and stable domain adaptation.}

\section{Experimental Results}
\label{sec:Results}

 Our implementation code is publicly available: \url{https://github.com/rostami-m/EventUDA}.

\subsection{Experimental Setup}

\textbf{Datasets:} We performed experiments  on the N-MNIST~\cite{orchard2015converting}, the  N-Caltech101\cite{https://doi.org/10.48550/arxiv.1507.07629}, and the CIFAR10-DVS\cite{lievent}  event datasets which are suitable for UDA. 

 \begin{itemize}
\item \textcolor{black}{ \textbf{N-MNIST}~\cite{orchard2015converting} is an event-based dataset that serves as the neuromorphic counterpart to the widely used MNIST handwritten digit classification benchmark. It was constructed by displaying the original static MNIST digits on a computer screen and recording the visual stimuli using a DAVIS dynamic vision sensor mounted on a motorized pan-tilt platform. 
Each event in the N-MNIST dataset encodes four attributes: the $x$ and $y$ pixel coordinates (in a $28 \times 28$ spatial grid), a polarity bit indicating whether the brightness increased or decreased, and a timestamp with microsecond resolution. The dataset consists of 60,000 training samples and 10,000 test samples, mirroring the original MNIST split. The typical duration of each event stream is approximately 300 milliseconds, which is commonly discretized into 300 time bins (1 ms per bin), resulting in a spike tensor of shape $[T \times 2 \times 28 \times 28]$, where $T = 300$ and the two channels represent ON and OFF events.
}

    \item \textcolor{black}{ \textbf{N-Caltech101}~\cite{https://doi.org/10.48550/arxiv.1507.07629} is an event-based classification dataset created as the neuromorphic counterpart to the well-known frame-based Caltech101 dataset~\cite{https://doi.org/10.48550/arxiv.1604.01518}. Each N-Caltech101 sample is generated by displaying a static Caltech101 image on a monitor and capturing it with an ATIS event camera mounted on a pan-tilt unit that mimics biological saccadic eye movements. This process converts static images into temporally asynchronous event streams, enabling spiking neural networks and event-driven algorithms to process them.
The event streams encode spatiotemporal information using four attributes: the $x$ and $y$ spatial coordinates, a polarity bit (indicating positive or negative brightness change), and a high-resolution timestamp (typically in microseconds). The native resolution of the ATIS sensor used for capturing is typically $240 \times 180$ pixels. This contrasts with the original Caltech101 images, which vary in size but are typically around $300 \times 200$ pixels and captured in RGB format.
Although the dataset originates from paired image-event recordings, our training setup follows UDA paradigm, where the RGB images and event data are treated as unpaired and event labels are withheld during training. This aligns with the UDA setting, where the goal is to transfer knowledge from a labeled source domain (Caltech101) to an unlabeled target domain (N-Caltech101). To maintain experimental consistency with prior work, we split both datasets into three subsets: 45\% for training, 30\% for validation, and 25\% for testing. Due to limited computational resources, we further reduced the size of the training set in our implementation.
 }

\item \textcolor{black}{  \textbf{CIFAR10-DVS} is the event-based counterpart of the standard CIFAR10 image dataset~\cite{Krizhevsky2009LearningML}. It comprises 10,000 event-stream recordings captured by a DVS camera with a spatial resolution of $128 \times 128$ pixels as CIFAR10 images are displayed on a monitor and undergo repeated closed-loop smooth (RCLS) movements~\cite{lievent}. The original CIFAR10 dataset consists of 60,000 color images ($32 \times 32$) in 10 classes, of which 1,000 images per class were selected and converted into the event modality~\cite{lievent}.
Although both datasets share the same object categories, the event-based domain differs significantly: CIFAR10-DVS encodes asynchronous spatiotemporal brightness changes rather than static color values. The event representations lack color and instead record intensity changes with microsecond timestamps. The modality and temporal characteristics amplify the domain gap, increasing the challenge of UDA for this dataset.
Another complicating factor is that CIFAR10 and CIFAR10-DVS are completely unpaired: there are no direct image-event sample correspondences to guide alignment during training. To our knowledge, this work is the first to attempt UDA from CIFAR10 to CIFAR10-DVS. We follow previous conventions by splitting both datasets into a 5:1 train–test split, aligning with practice in neuromorphic vision literature.
}

\end{itemize}

\textbf{Data Augmentation for Self-Supervised Learning:}
To calculate the self-supervised learning loss, we introduced a range of diverse augmentations on frame images, including color jitter, Gaussian blur, random resize, random affine, random crop, and random flip. Similarly, in the events domain, we applied random event shift, random flip, random resize, and random crop to generate augmented versions of the event data. By employing these diverse transformations, we aimed to enhance the generalizability of our model and increase its robustness in handling domain shifts within the input spaces~\cite{rostami2021cognitively}.
The purpose of these augmentations is twofold. First, they create a more comprehensive and varied training dataset, enabling the model to encounter a wider range of possible inputs during training. This exposure to diverse examples helps the model learn to adapt to various data distributions and prepares it to handle different types of inputs during inference.
Secondly, the augmentations act as a form of data regularization, preventing the model from overfitting to specific patterns or biases present in the original dataset.

\textbf{Architecture of Networks:} 
We used a modified version of ResNet18 \cite{https://doi.org/10.48550/arxiv.1512.03385}. Specifically, we utilized the first half of ResNet18, with a slight modification to the first Convolutional   layer, allowing for different input sizes, and the removal of max pooling layers. On the other hand, we retained the classifier as the second half of ResNet18. By doing so, we focus on aligning the distributions in the middle layer of ResNet18 during the training process.
To facilitate domain alignment, we introduced discriminator networks, which are simple  CNNs. These discriminator networks play a critical role in the adversarial learning process, where they aim to differentiate between features extracted from events and features extracted from frames.
  By incorporating these elements, we can achieve effective alignment of the content representations from events and frames.

\textbf{Optimization:}
We used the R-Adam  optimizer~\cite{https://doi.org/10.48550/arxiv.1908.03265}  with \begin{math}\beta_1 = 0\end{math}, \begin{math}\beta_2 = 0.999\end{math} with    \begin{math}lr = 10^{-4}\end{math}, and an exponential  decay rate of 0.95.  For Caltech101 and N-Caltech101, the batch size is 7 and for CIFAR10 and CIFAR10-DVS, the batch size is 21. 

\subsection{Comparative Results}

We present the performance of our approach on the two benchmarks and conducting a thorough comparison against existing State-of-the-Art (SOTA) event-based UDA methods.
In our reported results, we also include the outcomes of a second group of supervised learning algorithms that leverage transfer learning. Note that these supervised methods assume the availability of event labels during the training phase. It is natural to consider these algorithms  as upper bounds for UDA, providing a measure of evaluating the competitiveness for  UDA methods against situations at which we have enough event annotations.
All supervised methods applied to the CIFAR10-DVS dataset were specifically designed to create new models tailored to event data, utilizing various approaches for event representation. Additionally, we include a ``Baseline'' version of our approach, which undergoes training with all loss functions except self-supervised learning and uncorrelated conditioning. The Baseline model serves as a reference point for evaluating the impact of these specific components in our method which constitute  our novel contributions.
In our tables and throughout our analysis, we refer to our approach as ``Domain Adaptation for event data using self-supervised  learning and uncorrelated Conditioning'' (DAEC$^2$).

The comparison results  are presented in Tables \ref{tabl1},  \ref{tabl2}, and \ref{tab311}. The tables clearly indicate the superior performance achieved by our method compared to SOTA UDA algorithms.
Even the Baseline version of our approach demonstrates impressive improvements, with performance gains of  3.8\% on N-Caltech101 and 3.1\% on CIFAR10-DVS over the best UDA algorithm. However, we see that using the self-supervised learning and the uncorrelated conditioning losses lead to further improvement. These innovative methods contribute an additional 2.0\% and 3.7\% performance boosts, respectively, placing our approach in a competitive position even against supervised learning methods.
Notably, on the CIFAR10-DVS benchmark, our method outperforms the best supervised model, TDBN Resnet19 \cite{https://doi.org/10.48550/arxiv.2203.06145}, which was specifically designed to excel on event data. This observation is of paramount importance in practical scenarios as it highlights the potential of UDA in training models for event-related tasks without the need for data annotation.
Furthermore, it is essential to acknowledge the computational advantages of our approach. While E2VID and VID2E methods generate and utilize videos as the source domain, we work with images, making our approach computationally less expensive and more feasible for real-world applications.

 \begin{table}[ht]
\centering
\caption{{Classification accuracies on the  N-MNIST dataset.  Bold font denotes the best method in each  category of UDA and supervised learning   methods. For clarity, $\Delta$ denotes performance gain over the baseline with no adaptation.}}
\label{tab311}
\begin{tabular}{l|c|c}
\toprule
Method & Accuracy (\%) & $\Delta$ \\
\midrule
Baseline        & 41.80 & -     \\
SHOT \cite{liang2020we}          & 53.70 & +11.90 \\
Zhao et al. \cite{zhao2022transformer} & 53.20 & +11.40 \\
Liang et al.~  \cite{liang2021source}     & 68.80 & +27.00 \\
EventDance \cite{zheng2024eventdance} & 71.00 & \textbf{+29.20} \\
DAEC$^2$ & \textbf{90.89} & \textbf{+49.09} \\
\bottomrule
\end{tabular}
\end{table}

\begin{table}[h]
\begin{center}
 
\begin{tabular}{ c | c | c| c }      
\hline
Method&UDA&ResNet&Test Acc  \\  
\hline
 E2VID \cite{https://doi.org/10.48550/arxiv.1906.07165}&\checkmark & 34 & 82.1 \\ 
 VID2E \cite{https://doi.org/10.48550/arxiv.2110.10505}&\checkmark & 34& 80.7 \\  
 BG\cite{Messikommer20ral} & \checkmark&18 & 84.8  \\
 Baseline&\checkmark & 18 & 88.6  \\ 
 EventDance \cite{zheng2024eventdance} &\checkmark &  18& 66.77   \\ 
 EventDance \cite{zheng2024eventdance} &\checkmark &  34& 72.68   \\ 
SHOT \cite{liang2020we}    &\checkmark &  18& 44.26   \\ 
 SHOT \cite{liang2020we}   &\checkmark &  34& 61.12   \\ 
 Zhao et al. \cite{zhao2022transformer} &\checkmark &  18& 44.30  \\ 
 Zhao et al. \cite{zhao2022transformer} &\checkmark &  34& 61.10   \\ 
Liang et al.~  \cite{liang2021source} &\checkmark &  18& 49.33   \\ 
Liang et al.~  \cite{liang2021source} &\checkmark &  34& 64.54    \\

\textbf{DAEC$^2$}&\checkmark &  18& \textbf{90.6}  \\ 
 \hline
 E2VID\cite{https://doi.org/10.48550/arxiv.1906.07165}& $\times$ & 34 & 86.6  \\ 
 \textbf{VID2E}\cite{https://doi.org/10.48550/arxiv.2110.10505}&$\times$ &34 & \textbf{90.6}  \\ 
 EST\cite{https://doi.org/10.48550/arxiv.1904.08245}& $\times$& 34 & 81.7  \\ 
 HATS\cite{https://doi.org/10.48550/arxiv.1803.07913}& $\times$& 34 & 64.2  \\

 TDBN\cite{https://doi.org/10.48550/arxiv.2203.06145}& $\times$& 19 & 78.6  \\ 
\hline

\end{tabular}
\end{center}
\caption{Classification accuracies on the N-Caltech101 dataset. Bold font denotes the best method in each category of UDA and supervised learning methods.}
\label{tabl1}
\end{table}

\begin{table}[h]
\begin{center}
\begin{tabular}{ c | c | c| c }
\hline
Method &UDA & ResNet& Test Acc \\  
\hline
 BG(Our Repro.) & \checkmark& 18&  76.3 \\
 Baseline&\checkmark &  18&  79.4 \\ 
 EventDance \cite{zheng2024eventdance}&\checkmark &  18& 66.77  \\  
 SHOT \cite{liang2020we}   &\checkmark &  18& 36.97  \\  
 Zhao et al. \cite{zhao2022transformer}&\checkmark &  18& 36.42  \\  
 Liang et al.~  \cite{liang2021source}&\checkmark &  18& 37.26  \\  
   \textbf{DAEC$^2$}&\checkmark &  18& \textbf{83.1}  \\

 \hline
 DART\cite{https://doi.org/10.48550/arxiv.1710.10800}& $\times$& N/A & 65.8 \\
 Spike-based BP\cite{https://doi.org/10.48550/arxiv.2007.05785}& $\times$& N/A & 74.8 \\

 \textbf{TDBN}\cite{https://doi.org/10.48550/arxiv.2203.06145}& $\times$&  19& \textbf{78.0}  \\

\hline

\end{tabular}
\end{center}
\caption{Classification accuracies on the  CIFAR10-DVS dataset.  Bold font denotes the best method in each  category of UDA and supervised learning   methods. }
\label{tabl2}
\end{table}

\subsection{Ablative Experiments}

To demonstrate the significance of our major ideas in achieving improved performance, we conduct an ablative experiment. In Table~\ref{tabl3}, we present the results of our ablative study, focusing on the N-Caltech101 dataset. In this experiment, we explore four different scenarios, each involving the use of self-supervised learning and uncorrelated conditioning for model adaptation. The ``Baseline'' scenario represents the use of all loss functions, excluding the two specific loss terms.
By examining these different scenarios, we can assess the individual contributions of self-supervised learning and uncorrelated conditioning to the overall performance of our approach. This analysis provides valuable insights into the importance of each idea and its impact on the model's ability to adapt and perform well. Through this   analysis, we gain a deeper understanding of the role played by self-supervised learning and uncorrelated conditioning in aligning domain distributions.

\begin{table}[h]
\begin{center}


\begin{tabular}{ c |c} 
\hline
Condition&Validation Acc. \\ 
\hline
Baseline Without Event Attribute Encoder&80.7\\
Baseline&82.5\\
With Self-Supervised Learning Loss&83.6\\
With Uncorrelated Condition&83.7\\
With Both &84.4\\
\hline
\end{tabular}
\end{center}
\caption{Ablative experiments  on N-Caltech101 dataset.}
\label{tabl3}
\end{table}

We observe Table~\ref{tabl3}  that capturing how objects appear under event cameras is crucial for achieving optimal performance. Both of our novel ideas contribute   to improving the model's performance.
The first idea, focusing on extracting event-specific representations using self-supervised learning, proves to be   effective in enhancing the model's ability to understand and adapt to event-based data. By aligning the content representations between frames and events, self-supervised learning facilitates knowledge transfer and domain adaptation, leading to improved performance on event tasks.
Uncorrelated conditioning  also plays an effective role in obtaining better results. By enforcing the constraint that object features and event camera features are uncorrelated, the model can disentangle the diverse information contained in images and events. This leads to more robust and discriminative representations, which are vital for accurate knowledge transfer and improved performance.
Our experiments demonstrate that both of these ideas are complementary and contribute independently to the model's performance and when used together lead to the best performance.

\subsection{Analytical Experiments}

\subsubsection{Optimal Design} 
In our approach, we sought to discover the most effective design for our implementation by evaluating its performance on the validation set. To achieve this goal, we delved into exploring four distinct possibilities concerning the implementation of self-supervised learning and uncorrelated conditioning.
In these experiments, the primary focus of our investigation was to identify the ideal configuration for both self-supervised learning and uncorrelated conditioning.
Throughout the exploration process, we carefully assessed each of the four possibilities which allowed us to make well-informed decisions about the most suitable design for achieving the desired outcomes.

\begin{figure}[h]
    \centering
    \includegraphics[width=8cm, height=6.4cm]{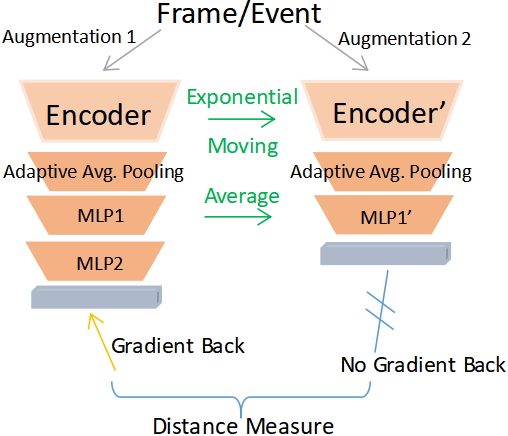}
    \caption{Momentum Encoder, labeled in $M_E$ in table \ref{tabl41}, if under average pooling, then there is no MLP1 and MLP1'.}
    \label{fig5}
\end{figure}

\begin{itemize}
 \item  Directly impose constraints on feature maps:   Utilizing $\ell_1$ loss facilitates the alignment of the two feature maps, thereby enforcing self-supervised  learning. However, implementing uncorrelated conditioning on these two feature maps is a more challenging task. As a result, we resorted to a simple and straightforward approach, employing negative $\ell_1$ loss in a naive manner.

\item Global average pooling on the feature maps: Following global average pooling, we obtain a vector representation. To implement the self-supervised learning loss, we experimented with two approaches: utilizing the $\ell_2$ distance and maximizing the cosine similarity. In Table 4, the negative cosine similarity ($-\cos(\cdot)$) is employed as the loss value due to convention.
Regarding uncorrelated conditioning, the most intuitive option is to use the cosine similarity metric. When the cosine similarity value between two vectors is zero, it indicates that they are uncorrelated. Therefore, we formulated the loss for this constraint as the absolute value of the cosine similarity to enforce uncorrelated conditioning.

\item Adding MLP after global  pooling then computing distances: explorations that performed in SimCLR\cite{https://doi.org/10.48550/arxiv.2002.05709}, have demonstrated that using MLP can be helpful. The constraints are implemented similarly to the previous case.

\item Inspired by BYOL~\cite{https://doi.org/10.48550/arxiv.2006.07733}, we also explored the implementation of momentum encoders and momentum MLP. The concept behind this approach, as depicted in Figure~\ref{fig5}, involves updating only one of the encoders when gradient backpropagation occurs, while the other encoder merely copies its parameters using an exponential moving average. Similarly, with momentum MLP, we use a setup where there is one encoder and two MLPs. During backpropagation, only one MLP gets updated, and the other MLP replicates its parameters through an exponential moving average.
Inspired by the BYOL method (Bring Your Own Latent) \cite{https://doi.org/10.48550/arxiv.2006.07733}, we discovered that it is not essential to employ negative samples. By eliminating the need for negative sampling, we were able to implement this condition effectively, even with our extremely limited computing resources. This approach streamlines the training process and further simplifies our implementation, making it more accessible and efficient for our specific computational setup.

\end{itemize}

\begin{table}
\begin{center}

\begin{tabular}{ c |c| c c|c c} 

\multicolumn{6}{ c }{Self-Supervised Learning Loss Alone} \\
\hline
&  No Proj.&\multicolumn{2}{ c| }{Avg. Pl.} & \multicolumn{2}{ c }{MLP} \\
\hline
Loss& \(\ell_1\)&\(-\cos\)&\(\ell_2\)&\(-\cos\)&\(\ell_2\) \\
\hline
Val. Acc&83.5&\textbf{83.6}&82.4&83.1&82.7 \\
\hline
\end{tabular}
\end{center}

\begin{center}
\begin{tabular}{ c |c| c |c} 

\multicolumn{4}{ c }{Uncorrelated Condition Alone} \\
\hline
&  No Proj.& Avg. Pl.&  MLP \\
\hline
Loss& \(-\ell_1\)&\(|\cos|\)&\(|\cos|\)\\
\hline
Val. Acc&82.2&\textbf{83.7}&82.6 \\
\hline
\end{tabular}
\end{center}

\begin{center}
\centering
\begin{tabular}{c|c|c|c}
\textbf{Val. Acc} & \textbf{Projection Type} & \textbf{Self-Supervised Loss} & \textbf{Notes} \\
\hline
\textbf{84.4} & Avg. Pl. & $-\cos$ & -- \\
83.6 & Avg. Pl. & $\ell_2$ & -- \\
82.7 & Avg. Pl. & $-\cos$ & $M_E$ \\
82.4 & Avg. Pl. & $\ell_2$ & $M_E$ \\
\hline
82.2 & MLP & $-\cos$ & -- \\
84.0 & MLP & $\ell_2$ & -- \\
82.9 & MLP & $-\cos$ & $M_E$ \\
83.3 & MLP & $\ell_2$ & $M_E$ \\
84.2 & MLP & $-\cos$ & $M_M$ \\
82.7 & MLP & $\ell_2$ & $M_M$ \\
82.8 & MLP & $-\cos$ & S \\
83.5 & MLP & $-\cos$ & $M_MS$ \\
\hline
83.5 & No Proj. & $\ell_1$ & UC loss = $-\ell_1$ \\
\end{tabular}
\caption{Effect of self-supervised learning loss design on validation accuracy for on N-Caltech101 dataset. Uncorrelated loss fixed as $|\cos|$ except for the last row.}
\label{tabl41}
\end{center}

\begin{center} 
\centering
\begin{tabular}{c|c|c|c}
\textbf{Val. Acc} & \textbf{Projection Type} & \textbf{UC Loss} & \textbf{Notes} \\
\hline
84.4 & Avg. Pl. & $|\cos|$ & -- \\
82.2 & MLP & $|\cos|$ & -- \\
82.8 & MLP & $|\cos|$ & S \\
83.5 & MLP & $|\cos|$ & $M_MS$ \\
-- & No Proj. & $-\ell_1$ & SSL loss = $\ell_1$ \\
\end{tabular}
\caption{Effect of uncorrelated conditioning loss design on validation accuracy for on N-Caltech101 dataset. SSL loss fixed as $-\cos$ except last row.}
\label{tabl42}
\end{center}


\end{table}

Tables \ref{tabl41} and \ref{tabl42} present the results obtained from our search for the optimal design, with $M_E$ denoting momentum encoders. When $M_E$ is listed under ``Avg. Pl.'', it means that there are no MLP1 and MLP1' components in Figure~\ref{fig5}. On the other hand, $M_M$ represents momentum MLP layers. For all MLPs, unless specified otherwise, the same MLP architecture is used for event content and frame content, while a different MLP is employed for event attribute. The abbreviation $S$ indicates that the same MLP is used for event content, frame content, and event attribute.
Analyzing the results from the table, we observe that both self-supervised learning loss and uncorrelated conditioning loss have positive contributions to the overall performance. However, our analysis leads us to the conclusion that the optimal design for achieving the best validation performance involves utilizing average pooling and applying the cosine similarity measure to both constraints. This configuration results in significant performance increases of 1.1\% and 1.2\%, respectively, when these constraints are used individually. Combining both constraints further boosts the performance by 1.9\%.

Interestingly, we also noticed that incorporating additional MLP layers, momentum encoders, and momentum MLP does not yield further performance improvement. We attribute this observation to the fact that introducing extra MLP layers allows the model more flexibility to circumvent the intended guidance provided by these two constraints. The primary purpose of these constraints is to steer the encoders towards extracting more informative features, and additional MLP layers seem to dilute this guidance.
Similarly, the lack of significant benefits from using momentum encoders can be explained by their inability to directly guide the encoders. In the case of BYOL \cite{https://doi.org/10.48550/arxiv.2006.07733}, momentum encoders prove helpful because the cosine similarity loss is the objective function itself, whereas in our scenario, the additional constraints serve as supplementary aids for the encoders.

In our pursuit of optimizing the model, we also conducted a grid search on the weights of the two losses to fine-tune them. We found out that both losses performed optimally when given equal weights, both set to one.
Overall, our   investigation and experimentation have led us to identify the most effective design choices for our implementation, achieving noteworthy performance gains and offering valuable insights into the impact of different components on the model's performance.

\subsubsection{Effect of UDA on Domain Alignment}
\begin{figure}[t!]
    \centering    
    \includegraphics[width=14cm, height=3.4cm]{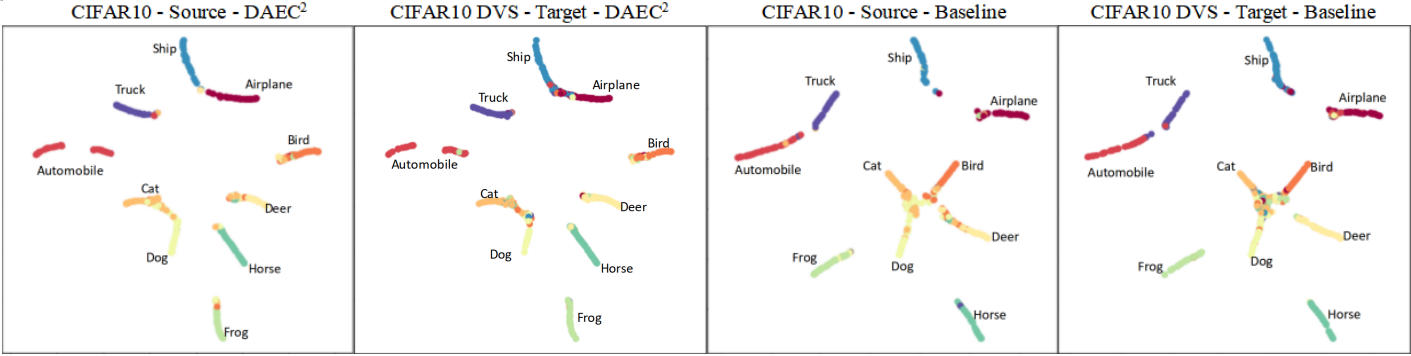}
        \caption{Effect of proposed method on data distribution in the shared embedding pace: we use UMap visualization for data representations at the model output layers using the test data of CIFAR10 and CIFAR10-DVS. }
         \label{fig8}
\end{figure}

To gain a better understanding of the impact of our algorithm on data representation, we conducted an empirical analysis and aimed to provide an intuitive justification for its effectiveness. To achieve this goal, we utilized the UMap visualization tool \cite{mcinnes2018umap}. UMap is a dimensionality reduction technique that allows us to project data representations from the latent embedding space into a lower-dimensional 2D space, enabling more accessible and intuitive visualizations.
In Figure \ref{fig8}, we present the UMap visualization of the output from the last layer of the classifier. The data used for this visualization was drawn from the test sets of both CIFAR10 and CIFAR10-DVS datasets. We compared the representations of the source and target domains after adaptation for both the baseline method and our proposed DAEC$^2$ algorithm.

As we had anticipated, the results show that both the baseline and DAEC$^2$ are effective in aligning the empirical distributions of the data representations from the source and target domains. However, a key observation is that when DAEC$^2$ is utilized, the class clusters are more separated and exhibit less overlap compared to the baseline. This finding indicates that our algorithm   improves the model's generalizability and enhances its ability to cope with domain shift in the input spaces.
The visual evidence provided by the UMap visualization strengthens our confidence in the effectiveness of DAEC$^2$ in refining the data representations, ultimately leading to a model that is better equipped to handle domain shift and achieve improved performance across different domains. The clearer separation of class clusters suggests that our algorithm effectively learns domain-invariant features, facilitating better transferability of knowledge between domains and resulting in a more robust and adaptable model.

\begin{figure} 
\includegraphics[width=12.1875cm, height=7.5cm]{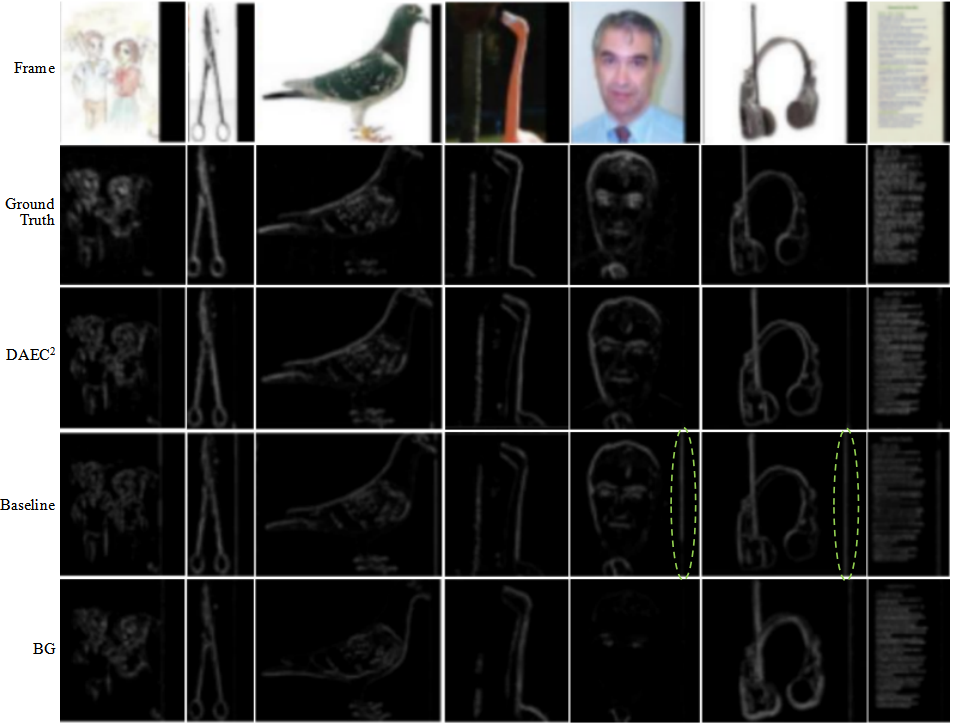}
\caption{Fake events images generated by three models: the input frame and event are drawn from test set, and they are not paired. A random event from test set is paired with a frame, which is not shown. Green circles indicate that unnecessary lines are present in events generated by baseline model but should be discarded. 
 More lines can be seen on other events generated by the baseline model. 
These lines are boundaries of input frames, and are successfully discarded when self-supervised  learning loss and uncorrelated conditioning are introduced.}
\label{fig7}
\end{figure}

\subsubsection{Effect of DAEC$^2$ on the quality of generated images:}
Figure \ref{fig7} showcases various examples of generated fake events produced by three different models:  DAEC$^2$, the baseline model, and BG\cite{Messikommer20ral}. We also visualize the our ground truth event images. The inputs for these examples are drawn from the test sets of Caltech101 and N-Caltech101 datasets, and it's important to note that these inputs are not paired, meaning they are unrelated images from different domains. Through a   visual examination, we can observe that our proposed DAEC$^2$ model achieves the highest visual quality among all the presented models.
In the absence of our proposed constraints, the baseline model fails to identify and discard all the useless features present in the generated events. These redundant features primarily include the boundaries between frame images and the padding, which are indicated by the green circles in Figure \ref{fig7}. Nearly every event generated by the baseline model exhibits such unnecessary lines, indicating that the model struggles to effectively remove irrelevant information.
However, by incorporating self-supervised learning loss and uncorrelated conditioning, our DAEC$^2$ model successfully identifies and filters out these unnecessary features from the input frames. As a result, the events generated by DAEC$^2$ appear much cleaner and exhibit fewer missing object boundaries, leading to a significant improvement in visual quality compared to the baseline.
Our proposed constraints play a crucial role in guiding the encoders to extract useful information while discarding irrelevant and redundant details from unpaired images. By effectively leveraging the self-supervised learning loss and uncorrelated conditioning, our model learns to focus on essential features, resulting in more accurate and visually appealing generated events.

\subsubsection{Effect of Self-Supervised Learning and Uncorrelated Conditioning losses on Learning Speed}

\begin{figure}[h]
\begin{center}
\begin{tikzpicture}
\begin{axis}[
    title={},
    xlabel={Epoch},
    ylabel={Test Acc (\%)},
    xmin=20, xmax=150,
    ymin=82, ymax=91,
    xtick={20,35,50,65,80,95,110,125,140},
    ytick={83,84,85,86,87,88,89,90,91},
    legend pos=south east,
    ymajorgrids=true,
    grid style=dashed,
]

\addplot[
    color=cyan,
    mark=circle,
    ]
    coordinates {
    (20,83.7)(24,84.1)(28,85.4)(32,84.2)(36,84.8)(40,85.9)(44,86.9)(48,87.2)(52,87.4)(56,85.8)(60,87.5)(64,87.3)(68,88.1)(72,87.4)(76,88.2)
    (80,88.0)(84,87.8)(88,88.2)(92,87.7)(96,87.8)(100,88.3)(104,88.0)(108,88.1)(112,87.4)(116,88.2)(120,88.5)(124,88.6)(128,88.1)(132,87.9)(136,87.7)(140,88.3)(144,87.7)(148,88.4)
    };
    \addlegendentry{Baeline}

\addplot[
    color=blue,
    mark=circle,
    ]
    coordinates {
    (20,82.1)(24,83.7)(28,85.5)(32,86.8)(36,86.5)(40,86.3)(44,87.2)(48,88.3)(52,87.3)(56,88.4)(60,87.2)(64,89.0)(68,88.4)(72,88.1)(76,88.6)
    (80,88.4)(84,88.3)(88,88.4)(92,89.5)(96,88.4)(100,88.9)(104,89.3)(108,88.8)(112,88.6)(116,88.5)(120,89.3)(124,89.5)(128,89.0)(132,90.2)(136,90.6)(140,90.0)(144,89.9)(148,90.5)
    };
    \addlegendentry{DAEC$^2$}
    
\end{axis}

\end{tikzpicture}
 
\caption{Test Accuracy versus training epochs on the NCaltech101 dataset}
 
\label{fig1a}

\end{center}
\end{figure}
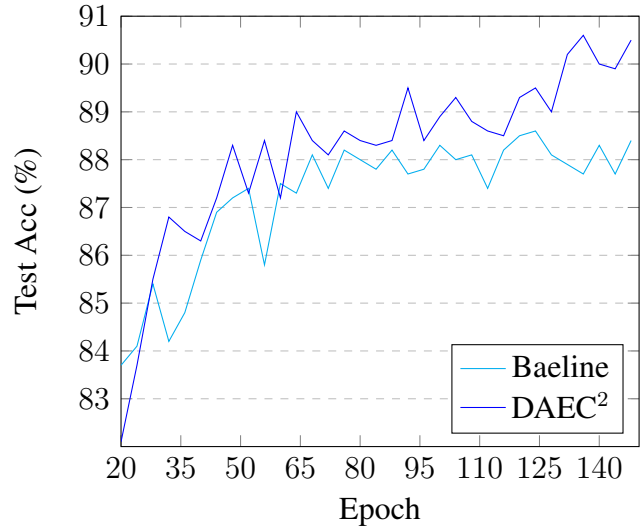

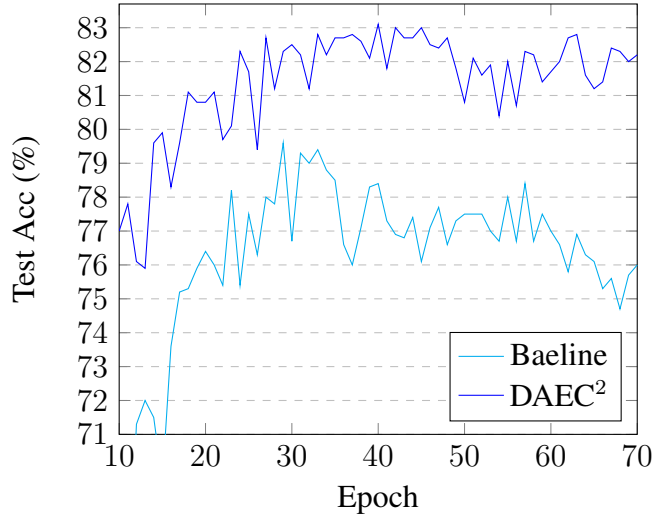
\begin{figure}[h]
\begin{center}
\begin{tikzpicture}
\begin{axis}[
    title={},
    xlabel={Epoch},
    ylabel={Test Acc (\%)},
    xmin=10, xmax=70,
    ymin=71, ymax=83.7,
    xtick={10,20,30,40,50,60,70},
    ytick={71,72,73,74,75,76,77,78,79,80,81,82,83},
    legend pos=south east,
    ymajorgrids=true,
    grid style=dashed,
]

\addplot[
    color=cyan,
    mark=circle,
    ]
    coordinates {
    (10,66.1)(11,65.8)(12,71.3)(13,72.0)(14,71.5)(15,69.6)(16,73.6)(17,75.2)(18,75.3)(19,75.9)(20,76.4)(21,76.0)(22,75.4)(23,78.2)(24,75.4)(25,77.5)(26,76.3)(27,78.0)(28,77.8)(29,79.6)(30,76.7)(31,79.3)(32,79.0)(33,79.4)(34,78.8)(35,78.5)(36,76.6)(37,76.0)(38,77.1)(39,78.3)(40,78.4)(41,77.3)(42,76.9)(43,76.8)(44,77.4)(45,76.1)(46,77.1)(47,77.7)(48,76.6)(49,77.3)(50,77.5)(51,77.5)(52,77.5)(53,77.0)(54,76.7)(55,78.0)(56,76.7)(57,78.4)(58,76.7)(59,77.5)(60,77.0)(61,76.6)(62,75.8)(63,76.9)(64,76.3)(65,76.1)(66,75.3)(67,75.6)(68,74.7)(69,75.7)(70,76.0)
    };
    \addlegendentry{Baeline}

\addplot[
    color=blue,
    mark=circle,
    ]
    coordinates {
    (10,77.0)(11,77.8)(12,76.1)(13,75.9)(14,79.6)(15,79.9)(16,78.3)(17,79.6)(18,81.1)(19,80.8)(20,80.8)(21,81.1)(22,79.7)(23,80.1)(24,82.3)(25,81.7)(26,79.4)(27,82.7)(28,81.2)(29,82.3)(30,82.5)(31,82.2)(32,81.2)(33,82.8)(34,82.2)(35,82.7)(36,82.7)(37,82.8)(38,82.6)(39,82.1)(40,83.1)(41,81.8)(42,83.0)(43,82.7)(44,82.7)(45,83.0)(46,82.5)(47,82.4)(48,82.7)(49,81.8)(50,80.8)(51,82.1)(52,81.6)(53,81.9)(54,80.4)(55,82.0)(56,80.7)(57,82.3)(58,82.2)(59,81.4)(60,81.7)(61,82.0)(62,82.7)(63,82.8)(64,81.6)(65,81.2)(66,81.4)(67,82.4)(68,82.3)(69,82.0)(70,82.2)
    };
    \addlegendentry{DAEC$^2$}
    
\end{axis}

\end{tikzpicture}
 
\caption{Test Accuracy versus training epochs on the CIFAR10-DVS dataset.}
\label{fig2a}
\end{center}
\end{figure}

Figure~\ref{fig1a} presents the learning curves of both our DAEC$^2$ model and the Baseline model, showcasing their performance on the N-Caltech 101 test set as training epochs progress. Similarly, Figure~\ref{fig2a} shows the learning curve specifically for the CIFAR10-DVS test set. A careful examination of these visualizations reveals  insights about the impact of incorporating self-supervised learning loss and uncorrelated conditioning in our DAEC$^2$ model.
First and foremost, we observe that the introduction of self-supervised learning loss and uncorrelated conditioning contributes to improved convergence speed during the training process. This observation means that our model is able to learn more efficiently and reach a desirable solution faster compared to the Baseline model.
Furthermore, not only does our DAEC$^2$ model converge faster, but it also achieves a better asymptotic solution. This observation indicates that our model performs consistently well even after convergence, demonstrating superior generalization capabilities. On the other hand, the Baseline model may suffer from performance degradation after reaching its peak accuracy.
Particularly on the CIFAR10-DVS test set, the Baseline model exhibits a decline in performance after attaining its peak accuracy. However, the incorporation of self-supervised learning loss and uncorrelated conditioning in our DAEC$^2$ model prevents such performance reduction from occurring. This observation emphasizes the importance of our proposed loss functions in ensuring the robustness of the model's performance across different datasets.
We  conclude that the inclusion of our loss functions   enhances the training process, leading to more robust and reliable models.

\begin{figure}[h]
\begin{center}
\begin{tikzpicture}
\begin{axis}[
    title={},
    xlabel={Training Hour (h) },
    ylabel={Test Acc (\%)},
    xmin=2, xmax=12,
    ymin=70, ymax=96,
    xtick={2,4,6,8,10,12},
    ytick={75,80,85,90,95},
    legend pos=south east,
    ymajorgrids=true,
    grid style=dashed,
]

\addplot[
    color=cyan,
    mark=square,
    ]
    coordinates {
    (2,89.4)(4,90.8)(6,91.9)(8,91.3)(10,91.3)(12,91.2)
    };
    \addlegendentry{N-Caltech101}

\addplot[
    color=blue,
    mark=square,
    ]
    coordinates {
    (2,90.0)(4,90.6)(6,91.1)(8,90.8)(10,90.9)(12,91.1)
    };
    \addlegendentry{Caltech101}
\addplot[
    color=red,
    mark=circle,
    ]
    coordinates {
    (2,77.1)(4,77.3)(6,80.1)(8,81.0)(10,80.4)(12,80.1)
    };
    \addlegendentry{CIFAR10-DVS}
\addplot[
    color=magenta,
    mark=circle,
    ]
    coordinates {
    (2,93)(4,94.3)(6,94.8)(8,94.9)(10,95.1)(12,94.7)
    };
    \addlegendentry{CIFAR10}
  
\end{axis}

\end{tikzpicture}
 
\caption{Comparison using simple supervised learning: horizontal axis shows the training hours.}
\label{lab1} 
\end{center}

\end{figure}
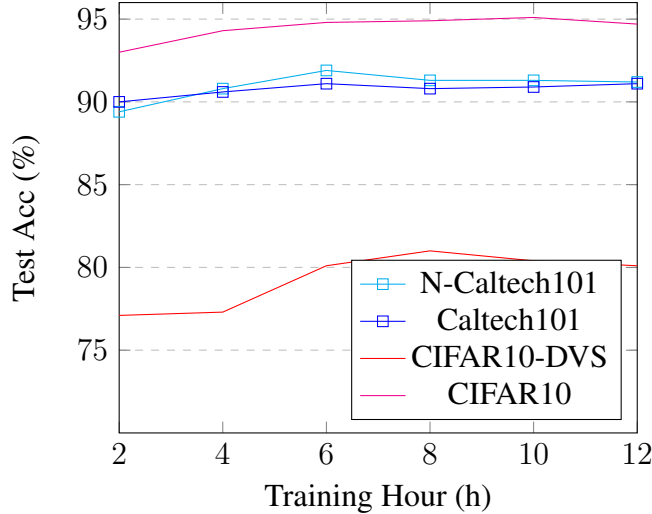


 \begin{figure}  
\begin{center}
\begin{tikzpicture}
\begin{axis}[
    title={},
    xlabel={Weight },
    ylabel={Val. Acc (\%)},
    xmin=0.5, xmax=2,
    ymin=81.9, ymax=84.5,
    xtick={0.5,0.75,1,1.25,1.5,2},
    ytick={82.5,83,83.5,84,84.5},
    legend pos=south east,
    ymajorgrids=true,
    grid style=dashed,
]

\addplot[
    color=blue,
    mark=square,
    ]
    coordinates {
    (0.5,83.3)(0.75,83.2)(1,84.4)(1.25,83.5)(1.5,83.7)(2,83.4)
    };
    \addlegendentry{Both Weights Same}

\addplot[
    color=green,
    mark=square,
    ]
    coordinates {
    (0.5,83.1)(0.75,83.1)(1,84.4)(1.25,83.9)(1.5,83.7)(2,83.8)
    };
    \addlegendentry{Self-Supervised  Learning Loss weight 1}
\addplot[
    color=black,
    mark=square,
    ]
    coordinates {
    (0.5,82.6)(0.75,83.0)(1,84.4)(1.25,83.1)(1.5,83.8)(2,84.1)
    };
    \addlegendentry{Uncorrelated Condition weight 1}
    
\end{axis}
\end{tikzpicture}
\end{center}
    
    \caption{Effect of loss weights on validation performance}
    \label{fig4a}
    
\end{figure}
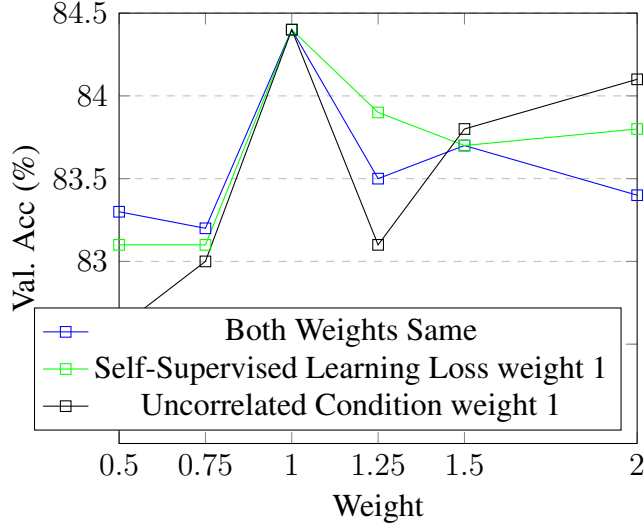

\begin{table}[h]
\label{tab1}
\begin{center}
\begin{tabular}{ c | c  }
\hline
 & Val. Acc (\%)  \\  
\hline
 \textbf{With everything} &  \textbf{84.4}  \\ 
  Without Fake Image Generation & 68.2  \\ 
 Without Refinement Net& 69.4 \\ 
 Without Event Discriminator& 72.0  \\ 
 Without Content Discriminator& 81.3  \\
\hline
\end{tabular}
\end{center}
\caption{Ablative experiments on the components of the proposed framework on N-Caltech101 validation set}
\end{table}

\subsubsection{Task Difficulty Comparison}

We conducted an experiment to compare the difficulty levels of two dataset pairs: (Caltech101, N-Caltech101) and (CIFAR10, CIFAR10-DVS). The results of this experiment are illustrated in Figure \ref{lab1}.
Upon analysis, we observed that the difficulty levels of Caltech101 and N-Caltech101 are quite similar, indicating that the adaptation from the source domain (Caltech101) to the target domain (N-Caltech101) is relatively manageable.
In contrast, the (CIFAR10, CIFAR10-DVS) pair presents a stark contrast, with CIFAR10-DVS proving to be a significantly more challenging dataset than CIFAR10. This observation suggests that the domain shift between these two datasets is more pronounced, making the adaptation between them more difficult.
However, the performances achieved by  the DAEC$^2$ model are very close to those of supervised learning with fully labeled data. For example, the supervised learning performances are 91.9\% and 81.0\% on N-Caltech101 and CIFAR10-DVS, respectively. In comparison, our DAEC$^2$ model achieves results that are only 1.3\% lower on N-Caltech101 and 2.1\% higher on CIFAR10-DVS, respectively.
This observation is particularly noteworthy because it demonstrates that our proposed novelties effectively bridge the gap between unsupervised methods and supervised learning. By narrowing the performance gap, our  model renders unsupervised learning approaches as competitive as traditional supervised learning methods. This development is significant as it instills optimism that deep learning techniques can be effectively employed in event-based problems without the burdensome necessity of annotating extensive event-based datasets.

\subsection{Performance Sensitivity with Respect to Hyperparameters}
To determine the optimal weights for the self-supervised learning loss and uncorrelated conditioning loss, we conducted a thorough grid search study using the N-Caltech101 validation set. The results of this study are graphically displayed in Figure~\ref{fig4a}.
During the grid search process, we   varied the weights assigned to the two loss functions, exploring different combinations to assess their impact on the model's performance. The objective was to find the weight configuration that would yield the highest overall performance on the validation set.

After   experimentation, we observed maintaining equal weights for both the self-supervised learning loss and uncorrelated conditioning loss consistently resulted in the best performance. Therefore, we found that assigning a weight of 1 to each of these loss functions is an effective strategy.
  By setting both weights to 1, we avoid the need for further fine-tuning and can achieve optimal results without any additional complexity in the loss function formulation.

\subsection{Computational Complexity}

\textcolor{black}{Experiments were conducted on NVIDIA H100-80 GB GPUs (CUDA 12.7, PyTorch 2.1). Unless otherwise stated, training used a mixed batch of 7 grayscale   images and 7 event histograms, totaling a batch size of 14 per device. We report information about the execution time in Table~\ref{tab7}. On a single GPU, a forward–backward pass takes approximately \textbf{9.8 minutes per epoch} (\(\sim0.23\) seconds per iteration over \(2 \times 2,500\) iterations). The model reaches stable classification performance (90\% top-1 accuracy on N-MNIST) within \textbf{30–40 epochs}, although we continue training up to 100 epochs to allow the adversarial and refinement components to converge more robustly. We also observe that scaling up to 4 GPUs with \texttt{torch.distributed} results in near-linear acceleration: the time per epoch drops to \textbf{2.6 minutes}, achieving an aggregate throughput of \textbf{1,610 images per second} (up from 425 on a single GPU), while maintaining similar convergence behavior (\(65 \pm 5\) epochs). Importantly, the per-GPU memory footprint remains constant.}

\textcolor{black}{
We have reported information on the number of trainable parameters of the architecture in Table \ref{tab8}. In terms of memory usage, 
the peak memory usage is \textbf{10.6 GB} per GPU, as reported by \texttt{nvidia-smi}. This includes \textbf{4.9 GB} for activations, \textbf{0.153 GB} for weights, and \textbf{0.306 GB} for optimizer states, along with approximately \textbf{5.2 GB} for CUDA kernels and miscellaneous buffers, consistent across single and multi-GPU settings.}

\begin{table}[h!]
    \centering
    \small
    \caption{Throughput and wall-clock convergence.}
    \label{tab7}
    \vspace{0.5em}
    \begin{tabular}{@{}lcccccc@{}}
        \toprule
        \textbf{GPU setup} & \textbf{Total batch} & \textbf{Time / epoch} & \textbf{Images s\(^{-1}\)} & \textbf{Conv. (epochs)} & \textbf{Wall-clock} \\
        \midrule
        1 \(\times\) H100-80 GB & 7 + 7 & 9.8 min & \(\approx 425\) & 65 \(\pm\)5 & \(\approx 10.7\) h \\
        4 \(\times\) H100-80 GB & 28 + 28 & 2.6 min & \(\approx 1,610\) & 65 \(\pm\)5 & \(\approx 2.9\) h \\
        \bottomrule
    \end{tabular}
\end{table}

\begin{table}[h!]
    \centering
    \caption{Trainable parameters per network component (millions).}
    \label{tab8}
    \vspace{0.5em}
    \begin{tabular}{@{}lc@{}}
        \toprule
        \textbf{Network component} & \textbf{Parameters (M)} \\
        \midrule
        Shared ResNet-18 trunk + two style encoders & 16.5 \\
        Task backend (residual + fc) & 11.0 \\
        Style decoder B (events) & 4.9 \\
        Cross-refinement B + discriminator & 4.2 \\
        Content discriminator & 1.6 \\
        \midrule
        \textbf{Total trainable} & \textbf{38.0} \\
        \bottomrule
    \end{tabular}
\end{table}

\section{Conclusions}
\label{sec:Conclusions}
We used self-supervised learning based on augmentation-invariant representation and uncorrelated conditioning, to tackle UDA from frame-based data to event-based data. Our experiments demonstrated the effectiveness of both of these ideas, leading to significant performance improvements on two event-based benchmarks. In fact, our approach yielded results that are comparable to those achieved by supervised learning methods that utilize annotated target domain data.
The idea of self-supervised learning based on augmentation-invariant representation proved to be a powerful approach in our UDA framework. By encouraging the model to learn features that are robust to various augmentations, we were able to improve the model's ability to generalize across different domains, resulting in enhanced performance on the event-based datasets.
Additionally,   uncorrelated conditioning  which involves conditioning on features extracted from the same input, emerged as an unprecedented idea in the context of UDA. The effectiveness of uncorrelated conditioning in our experiments has motivated us to consider its adoption in other tasks as part of our future work. This novel technique shows promise in scenarios where the extraction of uncorrelated features from the same input is essential or beneficial, such as improving object detection and other related tasks.
Given the compelling empirical evidence gathered through our experiments, we think that uncorrelated conditioning can find applications beyond UDA. Its potential to yield uncorrelated features from the same input opens up exciting possibilities for its utilization in various computer vision tasks and other domains.

 {
   
    \small
    \bibliographystyle{plain}
    \bibliography{ref}

\begin{thebibliography}{100}

\bibitem{alonso2019ev}
Inigo Alonso and Ana~C Murillo.
\newblock Ev-segnet: Semantic segmentation for event-based cameras.
\newblock In {\em Proceedings of the IEEE/CVF Conference on Computer Vision and
  Pattern Recognition Workshops}, pages 0--0, 2019.

\bibitem{baktashmotlagh2013unsupervised}
Mahsa Baktashmotlagh, Mehrtash~T Harandi, Brian~C Lovell, and Mathieu Salzmann.
\newblock Unsupervised domain adaptation by domain invariant projection.
\newblock In {\em Proceedings of the IEEE international conference on computer
  vision}, pages 769--776, 2013.

\bibitem{barrios2018movement}
Juan Barrios-Avil{\'e}s, Taras Iakymchuk, Jorge Samaniego, Leandro~D Medus, and
  Alfredo Rosado-Mu{\~n}oz.
\newblock Movement detection with event-based cameras: comparison with
  frame-based cameras in robot object tracking using powerlink communication.
\newblock {\em Electronics}, 7(11):304, 2018.

\bibitem{ben2006analysis}
Shai Ben-David, John Blitzer, Koby Crammer, and Fernando Pereira.
\newblock Analysis of representations for domain adaptation.
\newblock {\em Advances in neural information processing systems}, 19, 2006.

\bibitem{damodaran2018deepjdot}
Bharath Bhushan~Damodaran, Benjamin Kellenberger, R{\'e}mi Flamary, Devis Tuia,
  and Nicolas Courty.
\newblock Deepjdot: Deep joint distribution optimal transport for unsupervised
  domain adaptation.
\newblock In {\em Proceedings of the European Conference on Computer Vision
  (ECCV)}, pages 447--463, 2018.

\bibitem{bousmalis2017unsupervised}
Konstantinos Bousmalis, Nathan Silberman, David Dohan, Dumitru Erhan, and Dilip
  Krishnan.
\newblock Unsupervised pixel-level domain adaptation with generative
  adversarial networks.
\newblock In {\em Proceedings of the IEEE conference on computer vision and
  pattern recognition}, pages 3722--3731, 2017.

\bibitem{brock2019large}
Andrew Brock, Jeff Donahue, and Karen Simonyan.
\newblock Large scale gan training for high fidelity natural image synthesis,
  2019.

\bibitem{cao2018partial}
Zhangjie Cao, Lijia Ma, Mingsheng Long, and Jianmin Wang.
\newblock Partial adversarial domain adaptation.
\newblock In {\em Proceedings of the European conference on computer vision
  (ECCV)}, pages 135--150, 2018.

\bibitem{cao2019learning}
Zhangjie Cao, Kaichao You, Mingsheng Long, Jianmin Wang, and Qiang Yang.
\newblock Learning to transfer examples for partial domain adaptation.
\newblock In {\em Proceedings of the IEEE/CVF conference on computer vision and
  pattern recognition}, pages 2985--2994, 2019.

\bibitem{chen2020adversarial}
Minghao Chen, Shuai Zhao, Haifeng Liu, and Deng Cai.
\newblock Adversarial-learned loss for domain adaptation.
\newblock In {\em Proceedings of the AAAI conference on artificial
  intelligence}, volume~34, pages 3521--3528, 2020.

\bibitem{https://doi.org/10.48550/arxiv.2002.05709}
Ting Chen, Simon Kornblith, Mohammad Norouzi, and Geoffrey Hinton.
\newblock A simple framework for contrastive learning of visual
  representations, 2020.

\bibitem{chen2020big}
Ting Chen, Simon Kornblith, Kevin Swersky, Mohammad Norouzi, and Geoffrey
  Hinton.
\newblock Big self-supervised models are strong semi-supervised learners.
\newblock {\em arXiv preprint arXiv:2006.10029}, 2020.

\bibitem{chen2020mocov2}
Xinlei Chen, Haoqi Fan, Ross Girshick, and Kaiming He.
\newblock Improved baselines with momentum contrastive learning.
\newblock {\em arXiv preprint arXiv:2003.04297}, 2020.

\bibitem{choi2020learning}
Jonghyun Choi, Kuk-Jin Yoon, et~al.
\newblock Learning to super resolve intensity images from events.
\newblock In {\em Proceedings of the IEEE/CVF Conference on Computer Vision and
  Pattern Recognition}, pages 2768--2776, 2020.

\bibitem{courty2017optimal}
Nicolas Courty, R{\'e}mi Flamary, Devis Tuia, and Alain Rakotomamonjy.
\newblock Optimal transport for domain adaptation.
\newblock {\em IEEE Transactions on Pattern Analysis and Machine Intelligence},
  39(9):1853--1865, 2016.

\bibitem{doersch2017multi}
Carl Doersch and Andrew Zisserman.
\newblock Multi-task self-supervised visual learning.
\newblock In {\em Proceedings of the IEEE international conference on computer
  vision}, pages 2051--2060, 2017.

\bibitem{duan2012learning}
Lixin Duan, Dong Xu, and Ivor Tsang.
\newblock Learning with augmented features for heterogeneous domain adaptation.
\newblock {\em arXiv preprint arXiv:1206.4660}, 2012.

\bibitem{https://doi.org/10.48550/arxiv.2007.05785}
Wei Fang, Zhaofei Yu, Yanqi Chen, Timothee Masquelier, Tiejun Huang, and
  Yonghong Tian.
\newblock Incorporating learnable membrane time constant to enhance learning of
  spiking neural networks, 2020.

\bibitem{fang2022semi}
Zhen Fang, Jie Lu, Feng Liu, and Guangquan Zhang.
\newblock Semi-supervised heterogeneous domain adaptation: Theory and
  algorithms.
\newblock {\em IEEE Transactions on Pattern Analysis and Machine Intelligence},
  45(1):1087--1105, 2022.

\bibitem{gabourie2019learning}
Alexander~J Gabourie, Mohammad Rostami, Philip~E Pope, Soheil Kolouri, and
  Kuyngnam Kim.
\newblock Learning a domain-invariant embedding for unsupervised domain
  adaptation using class-conditioned distribution alignment.
\newblock In {\em 2019 57th Annual Allerton Conference on Communication,
  Control, and Computing (Allerton)}, pages 352--359, 2019.

\bibitem{gallego2020event}
Guillermo Gallego, Tobi Delbr{\"u}ck, Garrick Orchard, Chiara Bartolozzi, Brian
  Taba, Andrea Censi, Stefan Leutenegger, Andrew~J Davison, J{\"o}rg Conradt,
  Kostas Daniilidis, et~al.
\newblock Event-based vision: A survey.
\newblock {\em IEEE transactions on pattern analysis and machine intelligence},
  44(1):154--180, 2020.

\bibitem{Gallego2015EventbasedCP}
Guillermo Gallego, Christian Forster, Elias Mueggler, and Davide Scaramuzza.
\newblock Event-based camera pose tracking using a generative event model.
\newblock {\em ArXiv}, abs/1510.01972, 2015.

\bibitem{gehrig2020video}
Daniel Gehrig, Mathias Gehrig, Javier Hidalgo-Carri{\'o}, and Davide
  Scaramuzza.
\newblock Video to events: Recycling video datasets for event cameras.
\newblock In {\em Proceedings of the IEEE/CVF Conference on Computer Vision and
  Pattern Recognition}, pages 3586--3595, 2020.

\bibitem{https://doi.org/10.48550/arxiv.1904.08245}
Daniel Gehrig, Antonio Loquercio, Konstantinos~G. Derpanis, and Davide
  Scaramuzza.
\newblock End-to-end learning of representations for asynchronous event-based
  data, 2019.

\bibitem{ghifary2016deep}
Muhammad Ghifary, W~Bastiaan Kleijn, Mengjie Zhang, David Balduzzi, and Wen Li.
\newblock Deep reconstruction-classification networks for unsupervised domain
  adaptation.
\newblock In {\em European Conference on Computer Vision}, pages 597--613.
  Springer, 2016.

\bibitem{goodfellow2020generative}
Ian Goodfellow, Jean Pouget-Abadie, Mehdi Mirza, Bing Xu, David Warde-Farley,
  Sherjil Ozair, Aaron Courville, and Yoshua Bengio.
\newblock Generative adversarial networks.
\newblock {\em Communications of the ACM}, 63(11):139--144, 2020.

\bibitem{gouda2023improving}
Muhammed Gouda, Alessio Lugnan, Joni Dambre, Gerd van~den Branden, Christoph
  Posch, and Peter Bienstman.
\newblock Improving the classification accuracy in label-free flow cytometry
  using event-based vision and simple logistic regression.
\newblock {\em IEEE Journal of Selected Topics in Quantum Electronics}, 29(2:
  Optical Computing):1--8, 2023.

\bibitem{https://doi.org/10.48550/arxiv.2006.07733}
Jean-Bastien Grill, Florian Strub, Florent Altché, Corentin Tallec, Pierre~H.
  Richemond, Elena Buchatskaya, Carl Doersch, Bernardo~Avila Pires,
  Zhaohan~Daniel Guo, Mohammad~Gheshlaghi Azar, Bilal Piot, Koray Kavukcuoglu,
  Rémi Munos, and Michal Valko.
\newblock Bootstrap your own latent: A new approach to self-supervised
  learning, 2020.

\bibitem{https://doi.org/10.48550/arxiv.1512.03385}
Kaiming He, Xiangyu Zhang, Shaoqing Ren, and Jian Sun.
\newblock Deep residual learning for image recognition, 2015.

\bibitem{hu2020learning}
Yuhuang Hu, Tobi Delbruck, and Shih-Chii Liu.
\newblock Learning to exploit multiple vision modalities by using grafted
  networks.
\newblock In {\em European Conference on Computer Vision}, pages 85--101.
  Springer, 2020.

\bibitem{hu2021v2e}
Yuhuang Hu, Shih-Chii Liu, and Tobi Delbruck.
\newblock v2e: From video frames to realistic dvs events.
\newblock In {\em Proceedings of the IEEE/CVF Conference on Computer Vision and
  Pattern Recognition}, pages 1312--1321, 2021.

\bibitem{jian2023unsupervised}
Dayuan Jian and Mohammad Rostami.
\newblock Unsupervised domain adaptation for training event-based networks
  using contrastive learning and uncorrelated conditioning.
\newblock In {\em International Conference on Computer Vision}, 2023.

\bibitem{jolicoeurmartineau2018relativistic}
Alexia Jolicoeur-Martineau.
\newblock The relativistic discriminator: a key element missing from standard
  gan, 2018.

\bibitem{Krizhevsky2009LearningML}
Alex Krizhevsky.
\newblock Learning multiple layers of features from tiny images.
\newblock 2009.

\bibitem{le2019deep}
Tien-Nam Le, Amaury Habrard, and Marc Sebban.
\newblock Deep multi-wasserstein unsupervised domain adaptation.
\newblock {\em Pattern Recognition Letters}, 2019.

\bibitem{lee2019sliced}
Chen-Yu Lee, Tanmay Batra, Mohammad~Haris Baig, and Daniel Ulbricht.
\newblock Sliced wasserstein discrepancy for unsupervised domain adaptation.
\newblock In {\em Proceedings of the IEEE Conference on Computer Vision and
  Pattern Recognition}, pages 10285--10295, 2019.

\bibitem{lievent}
Cifar10-dvs Li.
\newblock an event-stream dataset for object classification, front.

\bibitem{li2020dual}
Lusi Li, Zhiqiang Wan, and Haibo He.
\newblock Dual alignment for partial domain adaptation.
\newblock {\em IEEE transactions on cybernetics}, 51(7):3404--3416, 2020.

\bibitem{li2022critical}
Shuang Li, Kaixiong Gong, Binhui Xie, Chi~Harold Liu, Weipeng Cao, and Song
  Tian.
\newblock Critical classes and samples discovering for partial domain
  adaptation.
\newblock {\em IEEE Transactions on Cybernetics}, 53(9):5641--5654, 2022.

\bibitem{https://doi.org/10.48550/arxiv.2203.06145}
Yuhang Li, Youngeun Kim, Hyoungseob Park, Tamar Geller, and Priyadarshini
  Panda.
\newblock Neuromorphic data augmentation for training spiking neural networks,
  2022.

\bibitem{liang2019exploring}
Jian Liang, Ran He, Zhenan Sun, and Tieniu Tan.
\newblock Exploring uncertainty in pseudo-label guided unsupervised domain
  adaptation.
\newblock {\em Pattern Recognition}, page 106996, 2019.

\bibitem{liang2020we}
Jian Liang, Dapeng Hu, and Jiashi Feng.
\newblock Do we really need to access the source data? source hypothesis
  transfer for unsupervised domain adaptation.
\newblock In {\em International Conference on Machine Learning}, pages
  6028--6039. PMLR, 2020.

\bibitem{liang2021source}
Jian Liang, Dapeng Hu, Yunbo Wang, Ran He, and Jiashi Feng.
\newblock Source data-absent unsupervised domain adaptation through hypothesis
  transfer and labeling transfer.
\newblock {\em IEEE Transactions on Pattern Analysis and Machine Intelligence},
  44(11):8602--8617, 2021.

\bibitem{lichtsteiner2008128}
Patrick Lichtsteiner, Christoph Posch, and Tobi Delbruck.
\newblock A 128$\backslash times 128 120 db 15 \backslash mu $ s latency
  asynchronous temporal contrast vision sensor.
\newblock {\em IEEE journal of solid-state circuits}, 43(2):566--576, 2008.

\bibitem{liu2020heterogeneous}
Feng Liu, Guangquan Zhang, and Jie Lu.
\newblock Heterogeneous domain adaptation: An unsupervised approach.
\newblock {\em IEEE transactions on neural networks and learning systems},
  31(12):5588--5602, 2020.

\bibitem{https://doi.org/10.48550/arxiv.1908.03265}
Liyuan Liu, Haoming Jiang, Pengcheng He, Weizhu Chen, Xiaodong Liu, Jianfeng
  Gao, and Jiawei Han.
\newblock On the variance of the adaptive learning rate and beyond, 2019.

\bibitem{long2015learning}
Mingsheng Long, Yue Cao, Jianmin Wang, and Michael Jordan.
\newblock Learning transferable features with deep adaptation networks.
\newblock In {\em Proceedings of International Conference on Machine Learning},
  pages 97--105, 2015.

\bibitem{long2018conditional}
Mingsheng Long, Zhangjie Cao, Jianmin Wang, and Michael~I Jordan.
\newblock Conditional adversarial domain adaptation.
\newblock In {\em Advances in Neural Information Processing Systems}, pages
  1640--1650, 2018.

\bibitem{long2017deep}
Mingsheng Long, Han Zhu, Jianmin Wang, and Michael~I Jordan.
\newblock Deep transfer learning with joint adaptation networks.
\newblock In {\em Proceedings of the 34th International Conference on Machine
  Learning-Volume 70}, pages 2208--2217. JMLR. org, 2017.

\bibitem{maqueda2018event}
Ana~I Maqueda, Antonio Loquercio, Guillermo Gallego, Narciso Garc{\'\i}a, and
  Davide Scaramuzza.
\newblock Event-based vision meets deep learning on steering prediction for
  self-driving cars.
\newblock In {\em Proceedings of the IEEE conference on computer vision and
  pattern recognition}, pages 5419--5427, 2018.

\bibitem{mcinnes2018umap}
Leland McInnes, John Healy, Nathaniel Saul, and Lukas Gro{\ss}berger.
\newblock {UMAP}: Uniform manifold approximation and projection.
\newblock {\em Journal of Open Source Software}, 3(29):861, 2018.

\bibitem{Messikommer20ral}
Nico Messikommer, Daniel Gehrig, Mathias Gehrig, and Davide Scaramuzza.
\newblock Bridging the gap between events and frames through unsupervised
  domain adaptation.
\newblock 2022.

\bibitem{morerio2017minimal}
Pietro Morerio, Jacopo Cavazza, and Vittorio Murino.
\newblock Minimal-entropy correlation alignment for unsupervised deep domain
  adaptation.
\newblock In {\em ICLR}, 2018.

\bibitem{morgenstern2014properties}
Yaniv Morgenstern, Mohammad Rostami, and Dale Purves.
\newblock Properties of artificial networks evolved to contend with natural
  spectra.
\newblock {\em Proceedings of the National Academy of Sciences},
  111(supplement\_3):10868--10872, 2014.

\bibitem{https://doi.org/10.48550/arxiv.2110.10505}
Manasi Muglikar, Diederik~Paul Moeys, and Davide Scaramuzza.
\newblock Event guided depth sensing.
\newblock 2021.

\bibitem{murez2018image}
Zak Murez, Soheil Kolouri, David Kriegman, Ravi Ramamoorthi, and Kyungnam Kim.
\newblock Image to image translation for domain adaptation.
\newblock In {\em IEEE Conference on Computer Vision and Pattern Recognition},
  pages 4500--4509, 2018.

\bibitem{https://doi.org/10.48550/arxiv.1507.07629}
Garrick Orchard, Ajinkya Jayawant, Gregory Cohen, and Nitish Thakor.
\newblock Converting static image datasets to spiking neuromorphic datasets
  using saccades, 2015.

\bibitem{orchard2015converting}
Garrick Orchard, Ajinkya Jayawant, Gregory~K Cohen, and Nitish Thakor.
\newblock Converting static image datasets to spiking neuromorphic datasets
  using saccades.
\newblock {\em Frontiers in neuroscience}, 9:437, 2015.

\bibitem{pan2019transferrable}
Yingwei Pan, Ting Yao, Yehao Li, Yu~Wang, Chong-Wah Ngo, and Tao Mei.
\newblock Transferrable prototypical networks for unsupervised domain
  adaptation.
\newblock In {\em Proceedings of the IEEE Conference on Computer Vision and
  Pattern Recognition}, pages 2239--2247, 2019.

\bibitem{peng2019moment}
Xingchao Peng, Qinxun Bai, Xide Xia, Zijun Huang, Kate Saenko, and Bo~Wang.
\newblock Moment matching for multi-source domain adaptation.
\newblock In {\em Proceedings of the IEEE/CVF international conference on
  computer vision}, pages 1406--1415, 2019.

\bibitem{planamente2021da4event}
Mirco Planamente, Chiara Plizzari, Marco Cannici, Marco Ciccone, Francesco
  Strada, Andrea Bottino, Matteo Matteucci, and Barbara Caputo.
\newblock Da4event: towards bridging the sim-to-real gap for event cameras
  using domain adaptation.
\newblock {\em IEEE Robotics and Automation Letters}, 6(4):6616--6623, 2021.

\bibitem{posch2014retinomorphic}
Christoph Posch, Teresa Serrano-Gotarredona, Bernabe Linares-Barranco, and Tobi
  Delbruck.
\newblock Retinomorphic event-based vision sensors: bioinspired cameras with
  spiking output.
\newblock {\em Proceedings of the IEEE}, 102(10):1470--1484, 2014.

\bibitem{https://doi.org/10.48550/arxiv.1710.10800}
Bharath Ramesh, Hong Yang, Garrick Orchard, Ngoc Anh~Le Thi, Shihao Zhang, and
  Cheng Xiang.
\newblock Dart: Distribution aware retinal transform for event-based cameras,
  2017.

\bibitem{rebecq2018esim}
Henri Rebecq, Daniel Gehrig, and Davide Scaramuzza.
\newblock Esim: an open event camera simulator.
\newblock In {\em Conference on robot learning}, pages 969--982. PMLR, 2018.

\bibitem{rebecq2019high}
Henri Rebecq, Ren{\'e} Ranftl, Vladlen Koltun, and Davide Scaramuzza.
\newblock High speed and high dynamic range video with an event camera.
\newblock {\em IEEE transactions on pattern analysis and machine intelligence},
  43(6):1964--1980, 2021.

\bibitem{https://doi.org/10.48550/arxiv.1906.07165}
Henri Rebecq, René Ranftl, Vladlen Koltun, and Davide Scaramuzza.
\newblock High speed and high dynamic range video with an event camera, 2019.

\bibitem{redko2017theoretical}
A.~Redko, I.and~Habrard and M.~Sebban.
\newblock Theoretical analysis of domain adaptation with optimal transport.
\newblock In {\em Joint European Conference on Machine Learning and Knowledge
  Discovery in Databases}, pages 737--753. Springer, 2017.

\bibitem{rostami2021lifelongww}
Mohammad Rostami.
\newblock Lifelong domain adaptation via consolidated internal distribution.
\newblock {\em Advances in neural information processing systems},
  34:11172--11183, 2021.

\bibitem{rostami2021transfer}
Mohammad Rostami.
\newblock {\em Transfer Learning Through Embedding Spaces}.
\newblock CRC Press, 2021.

\bibitem{rostami2022increasing}
Mohammad Rostami.
\newblock Increasing model generalizability for unsupervised visual domain
  adaptation.
\newblock In {\em Conference on Lifelong Learning Agents}, pages 281--293.
  PMLR, 2022.

\bibitem{rostami2023domain}
Mohammad Rostami, Digbalay Bose, Shrikanth Narayanan, and Aram Galstyan.
\newblock Domain adaptation for sentiment analysis using robust internal
  representations.
\newblock In {\em Findings of the Association for Computational Linguistics:
  EMNLP 2023}, pages 11484--11498, 2023.

\bibitem{rostami2021cognitively}
Mohammad Rostami and Aram Galstyan.
\newblock Cognitively inspired learning of incremental drifting concepts.
\newblock In {\em International Joint Conference on Artificial Intelligence},
  2023.

\bibitem{rostami2023overcoming}
Mohammad Rostami and Aram Galstyan.
\newblock Overcoming concept shift in domain-aware settings through
  consolidated internal distributions.
\newblock In {\em Proceedings of the AAAI conference on artificial
  intelligence}, volume~37, pages 9623--9631, 2023.

\bibitem{rostami2018crowdsourcing}
Mohammad Rostami, David Huber, and Tsai-Ching Lu.
\newblock A crowdsourcing triage algorithm for geopolitical event forecasting.
\newblock In {\em Proceedings of the 12th ACM Conference on Recommender
  Systems}, pages 377--381. ACM, 2018.

\bibitem{rostami2019deep}
Mohammad Rostami, Soheil Kolouri, Eric Eaton, and Kyungnam Kim.
\newblock Deep transfer learning for few-shot sar image classification.
\newblock {\em Remote Sensing}, 11(11):1374, 2019.

\bibitem{rostami2021lifelong}
Mohammad Rostami, Soheil Kolouri, Eric Eaton, and Kyungnam Kim.
\newblock Sar image classification using few-shot cross-domain transfer
  learning.
\newblock In {\em Proceedings of the IEEE/CVF Conference on Computer Vision and
  Pattern Recognition Workshops}, pages 0--0, 2019.

\bibitem{rostami2020generative}
Mohammad Rostami, Soheil Kolouri, Praveen Pilly, and James McClelland.
\newblock Generative continual concept learning.
\newblock In {\em Proceedings of the AAAI Conference on Artificial
  Intelligence}, volume~34, pages 5545--5552, 2020.

\bibitem{https://doi.org/10.48550/arxiv.1803.07913}
Amos Sironi, Manuele Brambilla, Nicolas Bourdis, Xavier Lagorce, and Ryad
  Benosman.
\newblock Hats: Histograms of averaged time surfaces for robust event-based
  object classification, 2018.

\bibitem{stan2021unsupervised}
Serban Stan and Mohammad Rostami.
\newblock Unsupervised model adaptation for continual semantic segmentation.
\newblock In {\em Proceedings of the AAAI Conference on Artificial
  Intelligence}, volume~35, pages 2593--2601, 2021.

\bibitem{stan2022domain}
Serban Stan and Mohammad Rostami.
\newblock Domain adaptation for the segmentation of confidential medical
  images.
\newblock In {\em Proceedings of the British Machine Vision Conference}, 2022.

\bibitem{stan2022secure}
Serban Stan and Mohammad Rostami.
\newblock Secure domain adaptation with multiple sources.
\newblock {\em Transactions on Machine Learning Research}, 2022.

\bibitem{sun2016deep}
Baochen Sun and Kate Saenko.
\newblock Deep coral: Correlation alignment for deep domain adaptation.
\newblock In {\em European conference on computer vision}, pages 443--450.
  Springer, 2016.

\bibitem{tian2021partial}
Yingjie Tian and Siyu Zhu.
\newblock Partial domain adaptation on semantic segmentation.
\newblock {\em IEEE Transactions on circuits and systems for video technology},
  32(6):3798--3809, 2021.

\bibitem{tzeng2017adversarial}
Eric Tzeng, Judy Hoffman, Kate Saenko, and Trevor Darrell.
\newblock Adversarial discriminative domain adaptation.
\newblock In {\em Proceedings of the IEEE Conference on Computer Vision and
  Pattern Recognition}, pages 7167--7176, 2017.

\bibitem{venkateswara2017deep}
Hemanth Venkateswara, Jose Eusebio, Shayok Chakraborty, and Sethuraman
  Panchanathan.
\newblock Deep hashing network for unsupervised domain adaptation.
\newblock In {\em Proceedings of the IEEE Conference on Computer Vision and
  Pattern Recognition}, pages 5018--5027, 2017.

\bibitem{wang2019event}
Lin Wang, Yo-Sung Ho, Kuk-Jin Yoon, et~al.
\newblock Event-based high dynamic range image and very high frame rate video
  generation using conditional generative adversarial networks.
\newblock In {\em Proceedings of the IEEE/CVF Conference on Computer Vision and
  Pattern Recognition}, pages 10081--10090, 2019.

\bibitem{wang2020eventsr}
Lin Wang, Tae-Kyun Kim, and Kuk-Jin Yoon.
\newblock Eventsr: From asynchronous events to image reconstruction,
  restoration, and super-resolution via end-to-end adversarial learning.
\newblock In {\em Proceedings of the IEEE/CVF Conference on Computer Vision and
  Pattern Recognition}, pages 8315--8325, 2020.

\bibitem{wang2018deep}
Mei Wang and Weihong Deng.
\newblock Deep visual domain adaptation: A survey.
\newblock {\em Neurocomputing}, 312:135--153, 2018.

\bibitem{wilson2020survey}
Garrett Wilson and Diane~J Cook.
\newblock A survey of unsupervised deep domain adaptation.
\newblock {\em ACM Transactions on Intelligent Systems and Technology (TIST)},
  11(5):1--46, 2020.

\bibitem{wu2023unsupervised}
Mengxi Wu and Mohammad Rostami.
\newblock Unsupervised domain adaptation for graph-structured data using
  class-conditional distribution alignment.
\newblock {\em arXiv preprint arXiv:2301.12361}, 2023.

\bibitem{https://doi.org/10.48550/arxiv.1604.01518}
Xinxing Xu, Joey~Tianyi Zhou, IvorW. Tsang, Zheng Qin, Rick Siow~Mong Goh, and
  Yong Liu.
\newblock Simple and efficient learning using privileged information, 2016.

\bibitem{yang2018learning}
Baoyao Yang, Andy~J Ma, and Pong~C Yuen.
\newblock Learning domain-shared group-sparse representation for unsupervised
  domain adaptation.
\newblock {\em Pattern Recognition}, 81:615--632, 2018.

\bibitem{zhai2019s4l}
Xiaohua Zhai, Avital Oliver, Alexander Kolesnikov, and Lucas Beyer.
\newblock S4l: Self-supervised semi-supervised learning.
\newblock In {\em Proceedings of the IEEE/CVF international conference on
  computer vision}, pages 1476--1485, 2019.

\bibitem{zhang2020learning}
Song Zhang, Yu~Zhang, Zhe Jiang, Dongqing Zou, Jimmy Ren, and Bin Zhou.
\newblock Learning to see in the dark with events.
\newblock In {\em European Conference on Computer Vision}, pages 666--682.
  Springer, 2020.

\bibitem{zhang2018collaborative}
Weichen Zhang, Wanli Ouyang, Wen Li, and Dong Xu.
\newblock Collaborative and adversarial network for unsupervised domain
  adaptation.
\newblock In {\em Proceedings of the IEEE conference on computer vision and
  pattern recognition}, pages 3801--3809, 2018.

\bibitem{zhao2022transformer}
Junwei Zhao, Shiliang Zhang, and Tiejun Huang.
\newblock Transformer-based domain adaptation for event data classification.
\newblock In {\em ICASSP 2022-2022 IEEE International Conference on Acoustics,
  Speech and Signal Processing (ICASSP)}, pages 4673--4677. IEEE, 2022.

\bibitem{zhao2022semantic}
Ying Zhao, Shuang Li, Rui Zhang, Chi~Harold Liu, Weipeng Cao, Xizhao Wang, and
  Song Tian.
\newblock Semantic correlation transfer for heterogeneous domain adaptation.
\newblock {\em IEEE Transactions on Neural Networks and Learning Systems},
  2022.

\bibitem{zheng2024eventdance}
Xu~Zheng and Lin Wang.
\newblock Eventdance: Unsupervised source-free cross-modal adaptation for
  event-based object recognition.
\newblock In {\em Proceedings of the IEEE/CVF Conference on Computer Vision and
  Pattern Recognition}, pages 17448--17458, 2024.

\bibitem{zhu2021eventgan}
Alex~Zihao Zhu, Ziyun Wang, Kaung Khant, and Kostas Daniilidis.
\newblock Eventgan: Leveraging large scale image datasets for event cameras.
\newblock In {\em 2021 IEEE International Conference on Computational
  Photography (ICCP)}, pages 1--11. IEEE, 2021.

\bibitem{CycleGAN2017}
Jun-Yan Zhu, Taesung Park, Phillip Isola, and Alexei~A Efros.
\newblock Unpaired image-to-image translation using cycle-consistent
  adversarial networks.
\newblock In {\em Computer Vision (ICCV), 2017 IEEE International Conference
  on}, 2017.

\end{thebibliography}
}

\end{document}